\documentclass[acmtog]{acmart} 
\makeatletter
\def\@ACM@checkaffil{
    \if@ACM@instpresent\else
    \ClassWarningNoLine{\@classname}{No institution present for an affiliation}%
    \fi
    \if@ACM@citypresent\else
    \ClassWarningNoLine{\@classname}{No city present for an affiliation}%
    \fi
    \if@ACM@countrypresent\else
        \ClassWarningNoLine{\@classname}{No country present for an affiliation}%
    \fi
}
\makeatother

\usepackage{enumitem}

\usepackage{booktabs, multirow} 
\usepackage{soul}
\usepackage{colortbl}
\usepackage{changepage,threeparttable} 
\usepackage{pifont}
\usepackage{adjustbox}
\usepackage{algorithm}
\usepackage{algpseudocode}
\usepackage{breakcites}

\usepackage{listings}
\usepackage{xcolor}
\definecolor{codegreen}{rgb}{0,0.6,0}
\definecolor{codegray}{rgb}{0.5,0.5,0.5}
\definecolor{codepurple}{rgb}{0.58,0,0.82}
\definecolor{backcolour}{rgb}{0.95,0.95,0.92}

\lstdefinestyle{mystyle}{
    backgroundcolor=\color{backcolour},   
    commentstyle=\color{codegreen},
    keywordstyle=\color{magenta},
    numberstyle=\tiny\color{codegray},
    stringstyle=\color{codepurple},
    basicstyle=\ttfamily\footnotesize,
    breakatwhitespace=false,         
    breaklines=true,                 
    captionpos=b,                    
    keepspaces=true,                 
    numbers=left,                    
    numbersep=5pt,                  
    showspaces=false,                
    showstringspaces=false,
    showtabs=false,                  
    tabsize=2
}
\newcommand{\longname}{$f$VDB}
\newcommand{\longnamespaced}{$f$VDB\ }

\newcommand{\eg}{\textit{e.g.}~}
\newcommand{\ie}{\textit{i.e.}~}

\newcommand{\JagT}{\texttt{JaggedTensor}\ }
\newcommand{\GridB}{\texttt{GridBatch}\ }

\newcommand{\GridBs}{\texttt{GridBatch}es\ }

\newcommand{\jdata}{\texttt{jdata}\ }
\newcommand{\jidx}{\texttt{jidx}\ }
\newcommand{\joffsets}{\texttt{joffsets}\ }

\newcommand*\CHECK{\ding{51}}

\AtBeginDocument{%
  }

\setcopyright{rightsretained}
\acmJournal{TOG}
\acmYear{2024} \acmVolume{43} \acmNumber{4} \acmArticle{133} \acmMonth{7} \acmDOI{10.1145/3658226}

\begin{document}

\title{\longname : A Deep-Learning Framework for Sparse, Large-Scale, and High-Performance Spatial Intelligence}

\author{Francis Williams}
\orcid{0000-0003-2189-881X}
\email{fwilliams@nvidia.com}
\affiliation{
  \institution{NVIDIA Research}
  \country{USA}
}

\author{Jiahui Huang}
\orcid{0000-0003-2189-881X}
\email{jiahuih@nvidia.com}
\affiliation{
  \institution{NVIDIA Research}
  \country{USA}
}

\author{Jonathan Swartz}
\orcid{0000-0001-6320-2636}
\email{jswartz@nvidia.com}
\affiliation{
  \institution{NVIDIA Research}
  \country{New Zealand}
}

\author{Gergely Kl\'{a}r}
\orcid{0000-0002-4569-5956}
\email{gklar@nvidia.com}
\affiliation{
  \institution{NVIDIA Research}
  \country{New Zealand}
}

\author{Vijay Thakkar}
\orcid{0000-0002-6158-6127}
\email{vithakkar@nvidia.com}
\affiliation{
  \institution{NVIDIA Research}
  \country{USA}
}

\author{Matthew Cong}
\orcid{0000-0003-2956-2050}
\email{mcong@nvidia.com}
\affiliation{
  \institution{NVIDIA Research}
\country{USA}
}

\author{Xuanchi Ren}
\orcid{0000-0001-6376-7100}
\email{xuanchir@nvidia.com}
\affiliation{
  \institution{NVIDIA Research}
  \country{Canada}
}

\author{Ruilong Li}
\orcid{0009-0005-7426-7650}
\email{ruilongl@nvidia.com}
\affiliation{
  \institution{NVIDIA Research}
  \country{USA}
}

\author{Clement Fuji-Tsang}
\orcid{0009-0002-0998-1581}
\email{cfujitsang@nvidia.com}
\affiliation{
  \institution{NVIDIA Research}
\country{Canada}
}
\author{Sanja Fidler}
\orcid{0000-0003-1040-3260}
\email{sfidler@nvidia.com}
\affiliation{
  \institution{NVIDIA Research}
  \country{Canada}
}

\author{Eftychios Sifakis}
\orcid{0000-0001-5608-3085}
\affiliation{
  \institution{University of Wisconsin-Madison}
  \country{USA}
}
\email{sifakis@cs.wisc.edu}
\affiliation{
\institution{NVIDIA Research}
  \country{USA}
}
\email{esifakis@nvidia.com}

\author{Ken Museth}
\orcid{0000-0002-9926-780X}
\email{kmuseth@nvidia.com}
\affiliation{
  \institution{NVIDIA Research}
  \country{USA}
}

\renewcommand{\shortauthors}{Williams et al.}
\begin{teaserfigure}
    \centering
    \includegraphics[width=\textwidth]{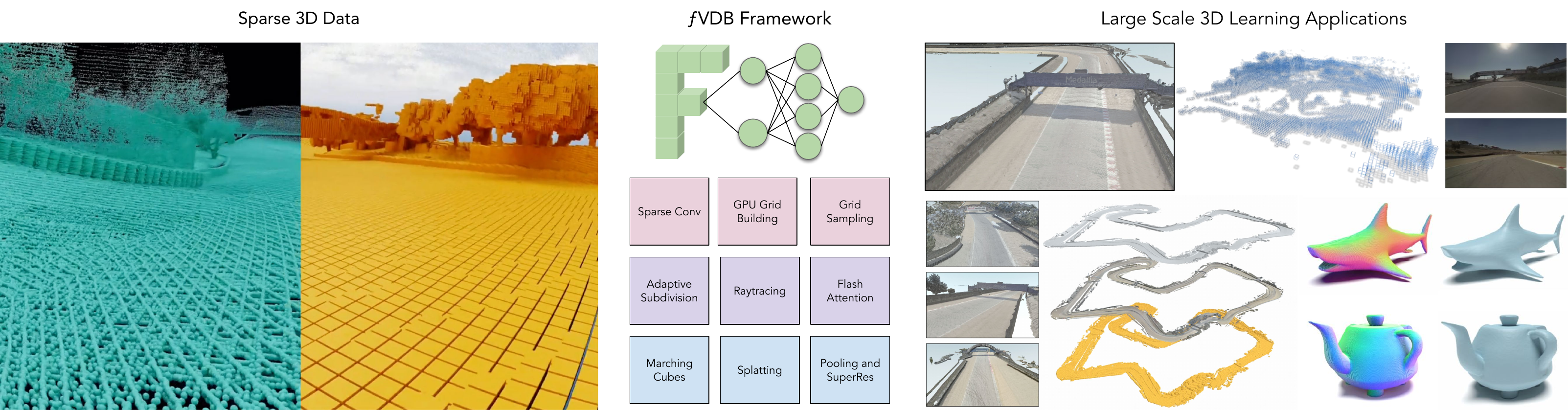}
    \caption{\longnamespaced is an integrated Deep Learning framework for large-scale, and high-performance spatial intelligence. It can process 3D data from a broad range of sources, including voxels, point clouds, and surface meshes. \longnamespaced also offers a rich set of state-of-the art differentiable operators, which can be used to build Deep Learning architectures for tasks in 3D Deep Learning, thus facilitating DL applications on large scale and high-resolution 3D data.}
    \label{fig:teaser}
\end{teaserfigure}

\begin{abstract}
We present \longname, a novel GPU-optimized framework for deep learning on large-scale 3D data. \longname\ provides a complete set of differentiable primitives to build deep learning architectures for common tasks in 3D learning such as convolution, pooling, attention, ray-tracing, meshing, etc. 
\longname\ simultaneously provides a \emph{much larger} feature set (primitives and operators) than established frameworks with no loss in efficiency: our operators match or exceed the performance of other frameworks with narrower scope. Furthermore, \longnamespaced can process datasets with much larger footprint and spatial resolution than prior works, while providing a competitive memory footprint on small inputs.  To achieve this combination of versatility and performance, \longname\ relies on a single novel VDB index grid acceleration structure paired with several key innovations including GPU accelerated sparse grid construction, convolution using tensorcores, fast ray tracing kernels using a Hierarchical Digital Differential Analyzer algorithm (HDDA), and jagged tensors. 
Our framework is fully integrated with PyTorch enabling interoperability with existing pipelines, and we demonstrate its effectiveness on a number of representative tasks such as large-scale point-cloud segmentation, high resolution 3D generative modeling, unbounded scale Neural Radiance Fields, and large-scale point cloud reconstruction. 
\end{abstract}


\maketitle

\section{Introduction}

Deep Learning methods have been foundational to solving a wide variety of previously intractable problems in computer science. These include building agents capable of passing the Turing test, generating high quality images from text prompts, speech and audio synthesis, and perception for robotics to name a few. Underlying these innovations lies a rich software ecosystem of deep learning primitives (such as convolution, pooling, and attention) which can be composed to build neural networks such as transformers or convolutional networks. These primitives are exposed to the programmer through deep learning frameworks such as PyTorch~\cite{paszke2019pytorch}, JAX~\cite{jax2018github}, or TensorFlow~\cite{tensorflow2015-whitepaper}. In common frameworks, these primitives operate on dense tensors of data, which often encode 1D or 2D signals (\eg text or images). In the case of tasks in 3D, dense tensors are fundamentally limited in size due to cubic scaling and memory constraints. Fortunately, 3D data is often sparse in nature, only requiring information to be encoded in a subset of the volume such as in the interior or near the surface of a shape. Thus, there has been an emergence of frameworks  \cite{choy20194d, tang2022torchsparse, tangandyang2023torchsparse} which operate on sparse 3D tensors of data. Correspondingly, many recent works propose network architectures which can operate on sparse 3D data \cite{choy20194d, ren2023xcube, qi2017pointnet, Wang_2017, huang2023nksr}. 

Past sparse 3D learning frameworks leverage hash tables as the primary data structure for mapping 3D integer coordinates to tensor data. Such a data structure works well for operators such as convolution and pooling, but the lack of spatial coherence of accesses makes it inefficient for operators such as sampling, splatting, and ray tracing without the use of auxiliary acceleration structures. Thus, past frameworks typically include a small number of operators such as convolution and pooling. However, we note that modern 3D learning tasks often involve a number of complex operators that must be combined together. For example, \cite{liu2023one2345} performs image-to-3d generation by unprojecting image features to a dense volume, leveraging a dense and sparse convolutional network to produce a sparse volume of learned features, then differentiably meshing and rendering this volume to produce a textured shape. Such a pipeline requires a number of complex differentiable operators (ray tracing, splatting, convolution, pooling, attention, meshing, and rendering) which can operate on sparse grids of learnable features. Currently, such pipelines are built using bespoke operators which glue together different acceleration structures (e.g. hash tables, occupancy bit fields and meshes) from different libraries. 

In this paper, we present \longname, a novel deep-learning framework for operating on sparse 3D tensors. Our framework provides a wide host of differentiable GPU accelerated 3D operators which can be easily composed to build complex 3D learning pipelines. Each of these operators delivers performance that is on par with or exceeding the performance of state-of-the-art operators from other frameworks which are much narrower in scope. Furthermore, \longnamespaced is memory efficient and is capable of processing \emph{much larger} inputs than existing alternatives. Table~\ref{tab:features} summarizes the features of \longnamespaced in contrast to existing 3D learning frameworks. 

The key innovation that enables us to develop a flexible and composable framework while still achieving state-of-the-art performance is a new data structure derived from NanoVDB \cite{Museth21:NanoVDB}, which we call \emph{IndexGrid}. This is paired with a novel ecosystem of tools for grid construction and traversal (see Section \ref{sec::GridBuild}), accelerated ray marching (see Section \ref{sec:hdda}),  and a novel data processing paradigm that unlocks aggressive optimizations in the application of stencil-based operators (\eg convolution in Section \ref{sec:sparse-conv}). 
While incorporating algorithms originally used in hash grid methods, which can be trivially adapted to our VDB structure, we also introduce new design paradigms that fit naturally within our representation. Specifically, we design optional convolutional alternatives that leverage efficient construction of locally densified, windowed views into the sparse data on which data regularity and aggressive utilization of tensorcores enable exceptional compute efficiency.

Our core contributions include:
\begin{itemize}
\item The design and deployment of a comprehensive API for spatial intelligence, with necessary primitives to accommodate a wide spectrum of high-value 3D Machine Learning tasks. 
\item A new sparse data structure, \emph{IndexGrid}, derived from NanoVDB \cite{Museth21:NanoVDB} but with a drastically re-imagined programming and execution model aimed to aggressively accelerate stencil-centric operations. 
\item A collection of GPU-optimized fast operators (convolution, attention, raytracing, etc) built around the IndexGrid structure, engineered to specifically target high efficiency on spatially sparse data.
\item A new benchmark for sparse convolution that highlights different workloads in terms of sparsity pattern and feature depth. 
\item Memory efficient algorithms which enable scaling to much larger inputs than prior works.
\item A demonstration of the applicability of our framework to a variety of end-to-end training and inference applications from a broad spectrum of 3D Deep Learning tasks.
\end{itemize}
\section{Related Work}

\paragraph{Sparse Voxel Data Structures for Deep Learning} Sparse 3D voxel grids are a common representation for deep learning on 3D data. Many past works such as \cite{choy20194d, spconv2022, tang2022torchsparse, tangandyang2023torchsparse} use a hash table to encode a mapping between 3D integer \verb|ijk| coordinates and offsets into a tensor of features. Such a mapping enables on average $O(1)$ lookup of arbitrary features, however accesses are not spatially coeherent. Furthermore, hash tables are not effective acceleration structures for operations such as ray marching since they are not a BVH. \par Another line of works~\cite{jatavallabhula2019kaolin, Wang_2017} use octrees instead of a hash table. These preserve spatial coherence and can be ray-marched efficiently, but at the cost of $O(\log N)$ access, and can grow quite deep for high resolutions. In contrast, \longnamespaced uses a fixed depth, shallow VDB~\cite{Museth13:VDB} tree, which enables $O(1)$ amortized reads and writes, and serves as an effective acceleration structure for a wide range of operations (See Table~\ref{tab:features}). VDB is a widely used data structure in computer graphics and simulation with several implementations including OpenVDB~\cite{Museth13:VDB} and NanoVDB~\cite{Museth21:NanoVDB} which implements a subset of OpenVDB on the GPU. More recently, NeuralVDB~\cite{kim2022neuralvdb} added neural compression on top of NanoVDB. 

Lastly, there are works that allow for the definition of custom, sparse volumetric data structures, such as the Taichi domain-specific language~\cite{hu2019taichi, hu2019difftaichi}, which provides a means to emit optimized, differentiable code, with emphasis on simulation tasks. In contrast, \longnamespaced is a general purpose framework targeting spatial sparsity, providing a collection of primitives that are useful to build end-to-end deep learning applications.

\paragraph{Deep Learning Frameworks} Deep learning architectures are constructed by composing together a series of differentiable operators with trainable parameters and optimizing those parameters via minimizing a loss functional over a dataset. In order to enable research and development of deep learning architectures, a number of software framework with composable primitives have arisen in the past decade. The most commonly used frameworks include PyTorch~\cite{paszke2019pytorch}, TensorFlow~\cite{tensorflow2015-whitepaper}, JAX~\cite{jax2018github}, and Keras~\cite{chollet2015keras}. These libraries expose primitives for operating on dense tensors of data (such as images and audio signals). 

\paragraph{3D Deep Learning Software} 3D deep learning tasks often involve more complex primitives which operate on sparse tensors. Common libraries such as the Minkowski Engine~\cite{choy20194d}, TorchSparse~\cite{tang2022torchsparse, tangandyang2023torchsparse}, and SpConv~\cite{spconv2022} add support for constructing sparse tensors with basic operations such as convolution and pooling. Other libraries such as NerfAcc~\cite{li2023nerfacc}, PyTorch3D~\cite{ravi2020pytorch3d} and Kaolin~\cite{jatavallabhula2019kaolin} provide other graphics operators such as ray tracing using dense bitfields and octrees as well as operators for meshes and graphs. Our framework, {\longname}  unifies many of these operations under a single library, providing a broader set of features than past works using only a single, highly versatile novel VDB acceleration structure. 

\paragraph{Applications of Sparse 3D Learning Frameworks} Frameworks for deep learning on sparse tensors have been used in a number of important applications in deep learning including Point Cloud Processing~\cite{Zhao_2021_ICCV, choy20194d}, 3D reconstruction of geometry from point clouds and/or images ~\cite{huang2023nksr, 10.1145/3550454.3555457, nerfstudio}, perception~\cite{choy20194d, 10160968, Shi_2020_CVPR}, and, more recently, 3D generative modelling~\cite{ren2023xcube}. \longnamespaced exposes the operators to perform all these tasks under a single library using only our IndexGrid VDB as an acceleration structure.  Section~\ref{sec:applications} shows some demonstrative applications of our framework to different tasks in 3D Deep Learning.

\section{Method}

As the name suggests, \longnamespaced is built on the VDB data structure \cite{Museth13:VDB}, which offers both compact storage and fast access to sparse 3D data. However, unlike previous adoptions of VDB, e.g.~in OpenVDB\cite{OpenVDB} and NanoVDB~\cite{Museth21:NanoVDB}, we have developed novel techniques specifically for machine learning on the GPU. This includes indexed storage, fast grid construction on the GPU, hierarchical Digital Differential Analyzers (DDAs) \cite{Museth14:HDDA}  for accelerated GPU raymarching, and blocked computation, each of which will be discussed below. Many of these improvements build on NanoVDB, yet they are essential to the \longname framework and play a critical role in enhancing the performance of our ML system.

\begin{figure*}
    \centering
    \includegraphics[width=0.88\linewidth]{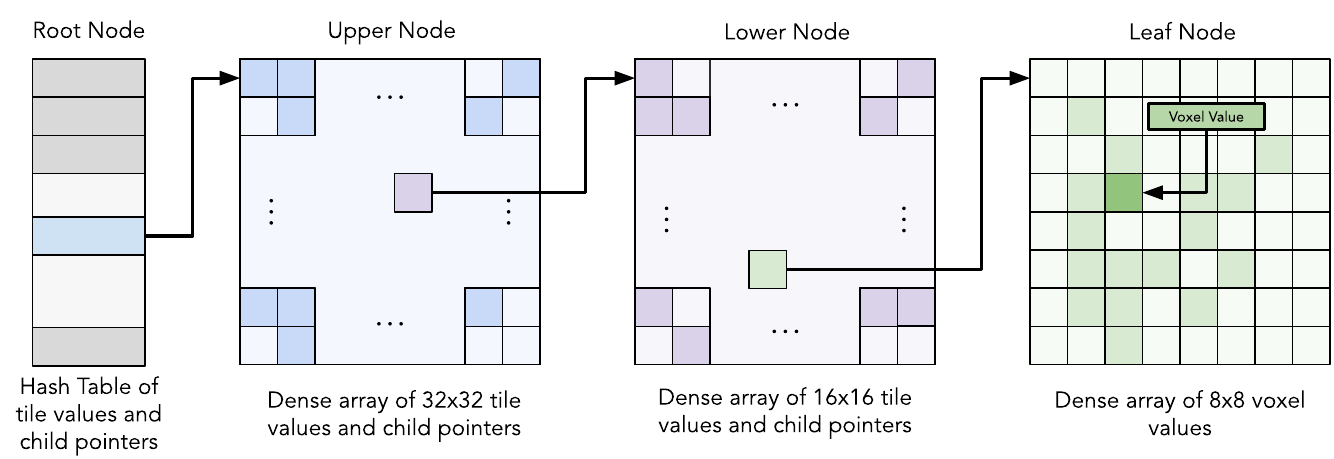}
    \caption{Illustration of a 2D VDB tree with the default configuration used in both OpenVDB and NanoVDB, which are 3D. The depth is 4, and the top-down fanout-factors are 32, 16, and 8 respectively. Values reside at all levels of the tree, and are denoted voxel values in the leaf node and tile values everywhere else.}
    \label{fig:2D_VDB}
\end{figure*}

\subsection{Background: VDB}

As a preamble, let's briefly summarize some of the main characteristics of the VDB data structure (see \cite{Museth13:VDB} for more details). At the core, VDB is a shallow 3D tree structure, with a hash table at the root level and a fixed hierarchy of dense child nodes with progressively decreasing block sizes. The default configuration in OpenVDB, and only configuration in NanoVDB, is three levels deep with the fan-out-factors 32, 16, and 8, i.e.~node sizes from root to leaf cover $4096^3$, $128^3$, and $8^3$ voxels respectively. This configuration is denoted $[\text{Map},5,4,3]$ in \cite{Museth13:VDB}, where the integers are $log2$ of the nodes fan-out-factors. The fact that VDB is shallow means that it supports fast random
(\ie coordinate-based) access to values. Furthermore, VDB allows for inverse tree-traversal, by means of node-caching, which in practice makes random-access $O(1)$. However, despite these attractive properties of VDB we found that it had several shortcomings when naively attempting to use it for ML applications on the GPU. Specifically, ML applications require more flexibility in terms of supporting complex high-dimensional data types, and the ML computations, e.g. sparse convolution, on the GPU are typically bandwidth-limited, which means random-access operations should be limited and data should be reused (cached) as much as possible.

\subsection{VDB IndexGrids for ML Features}

By design, standard VDB encodes data, e.g.~\verb|float| or \verb|Vec3f|, directly into the tree structure, i.e.~values and topology (\ie sparsity pattern) are mixed. That is, the data types (typically templated) and their numerical values are intertwined (both in terms of code and actual memory layouts) with their spatial occupancy (topology) information, compactly represented with bit-masks. This is problematic when dealing with data of arbitrary type and dimension (\ie ML features). It severely complicates code if each feature needs its own template specialization, and it is memory inefficient in cases when the sparsity (\ie topology) is shared between multiple feature/data types. Ironically, VDB was originally designed to handle situations where both topology and values are dynamic, but in ML we often found that topology is fixed, whereas data (payload) change in terms of type, value and dimension.\par
To overcome these inefficiencies we developed a completely new grid type in NanoVDB, dubbed IndexGrid, which effectively \emph{separates topology and values} encoded in VDB trees. Whereas the core idea behind IndexGrid is arguably simple, its efficient implementation is not. The idea is for the tree to return keys in the form of indices into external linear arrays of values as opposed to the data values, as is the case of standard VDB. In other words, the IndexGrid exclusively encodes topology information that is used to access any number of types of data values that resides in ``sidecars'', i.e. separate memory blocks. This seemingly trivial technique greatly simplifies code and allows for a single IndexGrid to be reused with multiple data (features), which amortizes the cost of encoding shared topology. \par There is another less obvious benefit to this IndexGrid, which is related to the fact that all nodes in VDB are fundamentally dense blocks, e.g.~a leaf node traditionally encodes $8^3=512$ values, regardless of the occupancy of the sparse data. A naive implementation of an IndexGrid indices all $512$ leaf values, but there is a much more memory efficient version of the IndexGrid that only indices the sparse (denoted active) leaf values. This significantly reduces the memory footprints of the sparse data (features stored externally as sidecars) since it eliminates the need to explicitly store values in leaf nodes that represent background values (as opposed to inserted active values). We achieve this sparse (vs dense) indexing of active values with the following highly efficient code.

\begin{lstlisting}[float, language=C++, caption=C++ code that computes sparse indices from coordinates.]
class LeafNode {uint64_t mOffset,mPrefixSum,mBitMask[8];
...
int off(int i,int j,int k){return (i&7)<<6|(j&7)<<3|k&7;}
uint64_t getValue(int i, int j, int k) {
    int m = this->off(i, j, k), n = m >> 6;
    uint64_t w = mBitMask[n], mask = 1 << (m & 63);
    if (w & mask == 0) return 0;// index to background
    uint64_t sum = n-- ? mPrefixSum >> (9*n) & 511 : 0;
    return sum + mOffset + countOn(w & (mask-1));
}};
\end{lstlisting}

\begin{lstlisting}[float, language=C++, caption=C++ code that computes offsets in nodes from coordinates.]
int lower::off(int i, int j, int k) {
    auto a = [](int n){return (n & 127) >> 3;};
    return a(i) << 8 | a(j) << 4 | a(k);// 0,1,..,16^3-1
};
int upper::off(int i, int j, int k) {
    auto a = [](int n){return (n & 4095) >> 7;};
    return a(i) << 10 | a(j) << 5 | a(k);// 0,1,..,32^3-1
};
\end{lstlisting}

In words, this compact code computes the linear offset from the signed coordinates \verb|i,j,k| to values stored in an external array, starting at \verb|mOffset|. Specifically, $m\in\{0,511\}$ is the linear index inside the leaf node, $n\in\{0,7\}$ is the offset into the 64-bit array \verb|mBitMask| that indicates which of the dense 512 values are active,\ie on. $w$ is the 64-bit word in \verb|mBitMask| that contains \verb|i,j,k|, and \verb|mask| masks out all higher bits in $w$, so as to only consider active states of values proceeding \verb|i,j,k|. Line 6 return a zero offset if \verb|i,j,k| maps to an inactive value, which corresponds to a unique background index. If $w$ is not the first word in \verb|mBitMask|, then line 7 extracts the preceding active value count encoded in the $7*9$ bits of \verb|mPrefixSum| as prefix sums of the first 7 64-bit words in \verb|mBitMask| (excluding last word). Finally line 8 computes the number of on bits in $w$, excluding any bits that comes after \verb|i,j,k|. \par Despite the apparent complexity of this code, it is very fast since it includes few (2) conditionals, and fast operations like bit and intrinsic function calls (e.g.~\verb|countOn|). Also, note that each leaf node in an IndexGrid only requires 80 bytes to encode all indices as opposed to over 4KB in \verb|nanovdb::LeafNode<uint64_t>|, i.e.~a memory reduction of over $50\times$ relative to a naive indexing approach. As mentioned above, IndexGrid also introduces memory saving by reusing topology for multiple data and avoiding explicitly storing inactive, i.e. background, values, which is especially important for sparse data.

\begin{figure}
    \centering
    \includegraphics[width=1.0\linewidth]{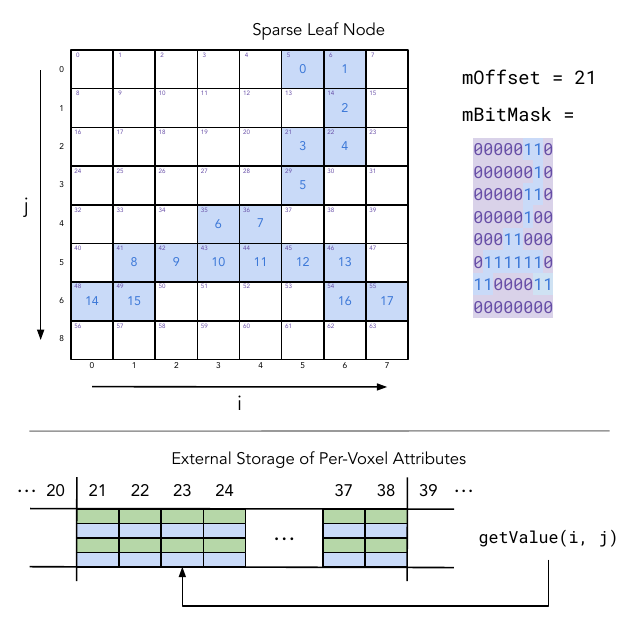}
    \caption{Illustration of dense local indexing (0-63) vs sparse global indexing (21-38) in a 2D leaf node of size $8^2=64$. The sparse global indexes correspond to offsets into a dense tensor of per-voxel attributes illustrated at the bottom as one column per attribute, allocated as sidecars to the IndexGrid.}
    \label{fig:IndexGrid}
\end{figure}

\subsection{GPU Accelerated IndexGrid Construction}\label{sec::GridBuild}
While shared topology information is efficiently handled with our new IndexGrid, there is still a need to dynamically change the sparsity patterns, \eg during morphological dilation, which is essential when building Level-of-Detail (LOD) hierarchies for sparse CNNs. In OpenVDB, dynamic topology is handled with allocation on insertion on the CPU, whereas in standard NanoVDB the topology is assumed to be fixed on both the GPU and CPU. Thus, there is a need to develop new techniques for building IndexGrids on the GPU, in order to rapidly build grids with different topology.

\begin{figure}
    \centering
    \includegraphics[width=0.88\linewidth]{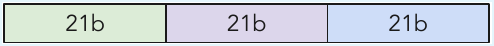}
    \caption{Breakdown of the 64 bit key constructed from voxel coordinates $i,j,k$ in step 2 of our build algorithm. The lower 21 bits (blue) encode the signed $k$ coordinate right-shifted $5+4+3=12$ bits, the next 21 bits (purple) encode the signed $j$ coordinate right-shifted 12 bits, and the upper 21 bits (green) encode the signed $i$ coordinate right-shifted by 12 bits. 
    }
    \label{fig:tile_key}
\end{figure}

\begin{figure}
    \centering
    \includegraphics[width=0.88\linewidth]{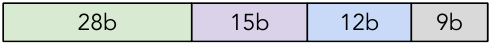}
    \caption{Breakdown of the unique 64 bit key constructed from voxel coordinates in step 5 of our build algorithm. The lower 9 bits (gray) encode the offset local into leaf nodes ($2^9=512=8^3$), the next 12 bits (blue) encode the local offsets into lower nodes ($2^{12}=4096=16^3$), the next 15 bits (purple) encode the local offsets into the upper nodes ($2^{15}=32768=32^3$), and finally the remaining upper 28 bits (green) encode tile ID (0 to total tile count - 1) into the hash table of the root node. Note, this imposes a limit of $2^{28}=268435456$ entries in the root hash table, which is extremely unlikely to be exceeded since each entry corresponds to a child node of the root that spans an index domain of size $4096^3$ voxels.}
    \label{fig:64bit_key}
\end{figure}

\begin{figure*}[ht]
    \centering
    \includegraphics[width=0.985\linewidth]{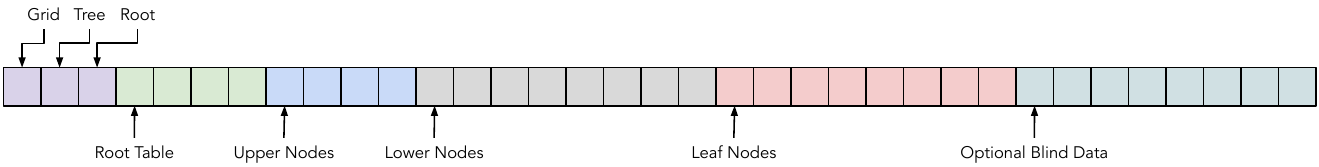}
    \caption{\textbf{NanoVDB Memory Layout:} By design NanoVDB, unlike OpenVDB, has the following serialized memory layout. Note that in our application the trailing optional blind data could be ML features of any type or dimension. Alternatively those data could reside in the separate memory block.}
    \label{fig:NanoVDB_layout}
\end{figure*}

A high-level description of our novel algorithm that builds IndexGrids from coordinates is as follows:

\begin{enumerate}[label*=\arabic*:]
    \item Input: $N$ signed voxel coordinates \verb|i,j,k|.
    \item Define $N$ 64-bit keys in Fig.~\ref{fig:tile_key}:\\
          \verb!a = k >> 12 | (j >> 12) << 21 | (i >> 12) << 42!
    \item Full radix sort of $N$ keys $a$.
    \item Run-Length-Encode $N$ keys $a$.
    \item For i,j,k in each run, $M=0,1,\ldots$, define keys in Fig.~\ref{fig:64bit_key}:\\
          \verb!b  = M << 36 | upper::off(i,j,k) << 21 |!\\
          \verb!     lower::off(i,j,k) << 9 | leaf::off(i,j,k)!
    \item Partial radix sort of keys, $b$, associated with run $M$.
    \item Upper node count is number of runs, $M$, in $a$.
    \item Lower node count is number of unique keys \verb|b>>21|.
    \item Leaf node count is number if unique keys \verb|b>>9|.
    \item Use node counts to allocate device memory in Fig~\ref{fig:NanoVDB_layout}.
    \item Build NanoGrid using the following top-down steps:
    \begin{enumerate}[label*=\arabic*:]
        \item Use $a$ to register upper nodes into the root table.
        \item Use \verb|b>>21| to register lower nodes into its parent nodes.
        \item Use \verb|b>>9| to register leaf nodes into its parent nodes.
        \item Use \verb|b&511| to register active voxels into \verb|leaf::mBitMask|.
    \end{enumerate}
    \item Optionally add ML features as blind data in Fig~\ref{fig:NanoVDB_layout}.
\end{enumerate}

Note that despite the complexity of the build algorithm outlined above, it is fast since virtually all steps can be performed in parallel on the GPU, and high-performance implementation of both radix sort and run-length-encoding are available in CUDA's CUB library\cite{merrill2015cub}. In fact, this build algorithm allow us to construct an IndexGrid from millions of voxel coordinates in a few milliseconds.

\subsection{Hierarchical DDA for fast Ray-Marching of VDB}\label{sec:hdda}

Efficient ray marching of our underlying data structure is essential for multiple tasks typical in 3D deep-learning workflows, including differentiable rendering, unprojecting image features into a 3D volume, depth computation, debug visualization, and final rendering. To this end we are using an acceleration technique, dubbed HDDA, that employs a hierarchy of Digital Differential Analyzers (DDAs), which accelerate ray marching on each of the tree levels of a VDB. While this technique was previously announced in a technical talk \cite{Museth14:HDDA}, we reiterate the process with more detail and technical elaboration in this paper.

\begin{figure*}[ht]
    \centering
    \includegraphics[width=0.8\linewidth]{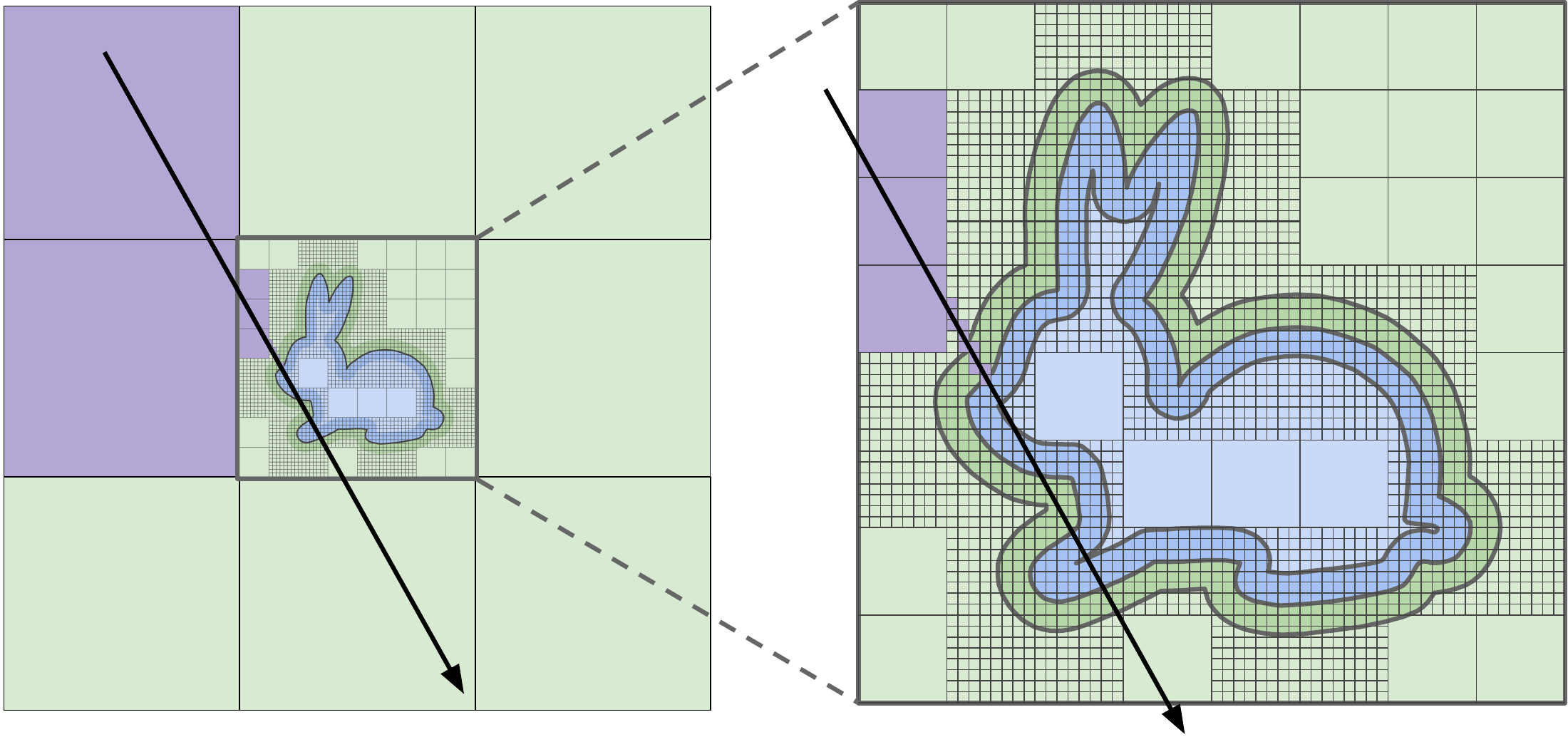}
    \caption{HDDA: Hierarchical Digital Differential Analyser allows for fast ray-marching of our VDB tree structure. It works by employing four DDAs, one per tree level, that facilitates efficient leapfrogging through empty space.  {\bf Left}: 2D illustration demonstrating the use of three DDAs to quickly skip large constant (or empty) regions of space, represented by upper tiles.
    {\bf Right}: 2D illustration of how two DDAs can accelerate ray-marching (violet squares represent tiles intersected by the ray).}
    \label{fig:HDDA}
\end{figure*}

The core idea of the HDDA is to associate four different DDAs with a given VDB tree structure -- one for each of the node levels corresponding to the coordinate domains $\{4096^3, 128^3, 8^3, 1^3\}$. In other words, the first DDA rasterizes a ray at the granularity of the root's child nodes of size $4096^3$ voxels, and the last (fourth) DDA rasterizes a ray at the fine voxel level. So, instead of slowly advancing the ray-marching at the voxel level, which would require numerous redundant random accesses into the VDB, we can use the coarser DDA in the hierarchy to effectively leapfrog through empty space. Given the fact that the VDB tree configuration is known at compile-time, we can use Template Meta-Programming to inline the logic of the four DDAs, resulting in a single high-performance HDDA. This significantly accelerates ray-marching and allows for real-time ray-tracing of VDB volumes on the GPU (typically marching millions of rays per second). We have illustrated this idea using two spatial dimensions in Fig.~\ref{fig:HDDA}. Our benchmark demonstrates a runtime that is 1.5x to 3x faster than DDA in the dense bitfield and over 100x less memory footprint, as reported in~\S~\ref{sec:exp:HDDA}.

\subsection{Accelerated Sparse Convolutional Operators}
\label{sec:sparse-conv}

\longnamespaced has been designed to be compatible with highly efficient algorithms for convolutional operations on sparse data, such as the Sorted Implicit Gemm (SpConv v2) paradigm used in TorchSparse++. We emphasize that leveraging such highly-tuned libraries in the context of our hierarchical, tree-based indexing structure is a straightforward exercise: \longnamespaced is effectively a locality-optimizing mapping between a sparse collection of lattice indices and a one-dimensional, linear index space. Contrary to random hash-based maps, \longnamespaced inherently provides the property that active indices that are geometrically proximate in the containing 3D lattice, will have high probability of also being proximate in linear index space. Conversely, active voxels corresponding to a contiguous sub-sequence of linear indices are highly likely to be geometrically clustered together in the containing 3D lattice. Other than this (favorable) inherent property of the \longnamespaced indexing scheme, our data structure is drop-in compatible with implementations that originate from hash-based structures (e.g. SpConv v2) by simply treating the linear index of each active voxel as a ``hash key'' (but with built-in locality properties). We have incorporated SpConv v2 into our operator toolkit and, as our micro-benchmarks reveal, we at minimum match the efficiency of TorchSparse++ at the operator level within our framework.

Even though SpConv v2 is trivially compatible with \longname, we have identified a number of scenarios where a new design perspective on convolutional kernel design can provide even higher performance. Although we present the circumstances leading to this acceleration opportunity, and detail our proposed algorithmic design choices, we highlight that \longnamespaced retains the ability to select the best applicable algorithm to match each case, including either the all-around performer SpConv v2, or our new kernels for those scenarios that warrant their use. Although we defer discussion of esoteric details of SpConv v2 to the related publications \cite{tangandyang2023torchsparse}, we highlight that its design is motivated by the following objectives: 
\begin{enumerate}[label=\alph*)]
\item Minimization of wasted computation, in the form of MACs (multiply-accumulate operations); relative to dense convolution, wasted computation could be either due to sparse occupancy of the background lattice, or sparse presence of the (max 27) stencil ``spokes'' across different lattice locations where a convolution stencil is applied. 
\item Maximization of regularity of operations; this typically manifests as an aspiration to perform the largest structured GEMM operation afforded by data layout and sparsity pattern. 
\item Minimization (or elimination) of scatter operations, and spatial localization of gather operations. 
\end{enumerate}
These design objectives become much more difficult to reconcile in the presence of significant sparsity and geometric irregularity. 

\paragraph{Scenario 1: Low-depth convolutions (\textbf{Leaf})} The first scenario where approaches striving for economy of computation might face diminishing returns is when the kernel is severely memory-bound. This possibility can easily materialize in the case of a convolution where both the input and output feature dimension is relatively low (e.g. not exceeding 8-16). As a tangible example: consider a convolution at TF32/FP32 precision with activation dimension of 8, and output dimension of 16. A \emph{dense} convolution at those depths, applied to an $N^3$ grid requires streaming at minimum $96N^3$ bytes (assuming perfect caching), and the performance of $6912N^3$ operations. On an RTX 6000 Ada Generation GPU (peak memory bandwidth of $960GB/s$) this would require about $70$TFLOP/s, which is an achievable compute density, to have this kernel be memory- rather than compute-bound. The calculus is not so straightforward when we contemplate sparsity, but we have practically witnessed this operation being pronouncedly memory-bound even at (local) sparsity of as little as 15-20\%. This is due to the inefficiency of necessary gather operations, the cost of indirection for accessing low-depth feature vectors, and the overhead of indexing data structures themselves. Additionally, even compute efficiency may be challenging due to the complexity of harvesting large-enough GEMM operations when the contraction dimension (8, in this example) is so shallow. 

In light of this, we consider an alternative where we prioritize regularity over sparsity of computation, essentially tolerating a higher compute burden for the sake of more local structure. Specifically, we have implemented a kernel that performs \emph{local densification} in GPU shared memory, at the level of an $8\times 8\times 8$ \longnamespaced leaf node, and performs a fully regular and (locally) dense convolution within this window. In detail, we allocate space in shared memory for a locally densified copy of the input activations in a window of size $10\times 10\times 10$ stradding the leaf node, plus a one-voxel halo in its immediate neighborhood (a footprint of 31.25KB for 4-byte FP32/TF32 data, at feature width of 8). Likewise, the output of this operation is an $8\times 8\times 8$ buffer of 16-wide output feature vectors (footprint of 32KB) also stored in shared memory. We subdivide the $8^3$ local domain into 32 $8\times 2 \times 1$ subtiles, assign each of them to a warp (1024 total threads) and use $16\times 16 \times 8$ WMMA tensorcore GEMMs (at TF32 precision with FP32 accumulate) within each warp to apply each of the 27 spokes of the stencil. Even though this paradigm clearly performs more computation than strictly necessary (foregoing sparity due to either voxel or stencil occupancy), the regularity of the computation in combination with the memory-bound nature of this scenario allows for superior performance (relative to our SpConv v2 default backend) in leaf nodes that have an occupancy of 20\% or higher (all the way to an approximately 2.5x-3x advantage for a dense domain). It should also be noted that no auxiliary indexing structures are necessary for this kernel approach, all gather offsets are computed directly and efficiency  from the (very lightweight) metadata of the core \longnamespaced tree structure, taking advantage of amortization. Finally, due to the compact and local storage of all (output) feature vectors within a leaf node, the writeback of the convolution result into global memory occurs on a fully sequential memory range (all active indices within a leaf node are sequentially indexed). 

\paragraph{Scenario 2: High local occupancy convolutions (\textbf{Brick})} The second scenario we target for a tuned approach is when the sparsity pattern exhibits high density in the vicinity of active indices (e.g. when on average every active index has more than 70-80\% of its stencil neighbors as active), even though the domain is macroscopically sparse. Typical cases where this scenario materializes is when the active indices are predominantly clustered in a narrow band of small but nontrivial thickness (e.g. 2-3 voxels wide), and also on dense or semi-dense domains that are still targeted with our \longnamespaced representation. In addition, we look for instances where such topology is coincident with moderate-to-high depth of input/output features (width of 32 or higher), when the kernel no longer is memory-bound as in scenario 1 above. For this case, we have implemented a solution that replicates the local densification paradigm, as above, but instead of this being performed at the granularity of an $8\times 8\times 8$ window, we focus on a kernel that monolithically produces the convolution output on a narrower $4\times 2\times 2$ window. Input activations are fetched on-demand from the spatial extent encompassing the $6\times 4\times 4$ window (including a 1-voxel halo) around the $4\times 2\times 2$ block. We have developed a custom tensorcore implementation of the convolution operation using the CuTe library \cite{Thakkar_CUTLASS_2023} that achieves exceptionally high compute density (exceeding 70\% peak compute bandwidth for moderate feature depths of about 32-64, and reaching above 90\% for feature depths of 128 or higher) for the task of computing the locally-dense convolution on the $4\times 2\times 2$ output window. Any residual suboptimality in this case is due to inactive voxels at the scale of the $4\times 2\times 2$ window, or stencil spokes that are not present for any of the active voxels. In practice, we have observed that for occupancy patterns that exceed 60-70\% on average across such windows, this implementation outperforms SpConv v2, with the most notable margin observed in dense or semi-dense domains that have even higher average occupancy. 

\paragraph{Scenario 3: Highly sparse topology, high feature depth (\textbf{LGGS})} The last scenario where we have provided a custom implementation addresses the instance where the occupancy pattern is so sparse that on average every active index is expected to have no more than 4-5 active neighbors (out of 26 max). In addition, this has to be combined with relatively high feature depth, typically of 128 or above. This scenario is characteristic of LiDAR data, as those presented in SemanticKITTI \cite{behley2021ijrr, geiger2012cvpr}. Although our default SpConv v2 implementation performs an adequate job at minimizing wasted MAC operations, the number of those may still exceed the essential MACs mandated by the stencil occupancy of active indices. 

In principle, if our sole objective was to minimize wasted MACs, the traditional gather-GEMM-scatter paradigm provides a pathway to achieving this goal. However, the reasons why the straightforward implementation of this paradigm will typically underperform SpConv v2 is due to the need for several independent streaming passes over the input activations (one for each of the 27 stencil offsets), and due to the suboptimality of scattering results to global memory. We circumvent these concerns by taking the following steps: 
\par(a) We block the gather-GEMM-scatter operation so that it is performed on a contiguous subsequence of output indices from the \longnamespaced data structure, typically 64 indices at a time. Due to the locality of the \longnamespaced mapping, those indices are expected to correspond to highly clustered geometric coordinates from one or more IndexGrid leaf nodes. 
\par(b) Instead of scattering results to global memory, we use a temporary buffer in GPU shared memory as the destination of scatter operations on these 64 indices, which collect the contribution of each of the 27 stencil offsets within this block. At the end of the local computation, this result is sequentially copied back to global memory without the need of a scatter operation. 
\par(c) For each of the 27 stencil offsets, we collect all input/output index pairs that are linked by this offset (such that the output index is within the range of the block being processed), and pack them contiguously again in shared memory buffers. For each stencil offset, the input of this packed buffer is gathered from global memory (benefiting from locality across offsets). A GEMM operation is performed to produce the output, still in packed format, to be scattered (purely in shared memory) to the accumulation buffer that stores all 64 output vectors. We pad these packed collections of input/output index pairs to the next multiple of 16, for purposes of easy mapping to tensorcore-accelerated GEMM. This is the only source of wasted MACs, which is now limited to at most 15 MACs per block of 64 output indices (practically, the expected length of this padding is closer to 8 entries per 64 output indices). 

Our benchmarks demonstrate a runtime that is approximately 25\% faster than SpConv v2 (at feature length 128 or higher) for the single-scan point clouds of SemanticKITTI.

\subsection{\longnamespaced Framework Overview}

At its core, \longnamespaced exposes a set of differentiable deep learning primitives which operate a \emph{minibatch} of sparse voxel grids. \ie a set of multiple sparse voxel grids where each voxel contains some multi-dimensional tensor of data. 
To encode such a minibatch of grids, \longnamespaced employs two classes: a \GridB which represents a set of NanoVDB index grids (one per item in the batch) and a \JagT which encodes a tensor of per-voxel features at each voxel in the minibatch. 
Internally, a \GridB is simply a contiguous block of NanoVDB IndexGrids stored one after the other with some metadata to quickly access any grid in the batch. Below, we give a description of the \GridB and \JagT classes as well as a summary of the primary operators exposed to the programmer by \longnamespaced.

\subsubsection{\JagT}
\begin{figure*}
    \centering
    \includegraphics[width=0.99\linewidth]{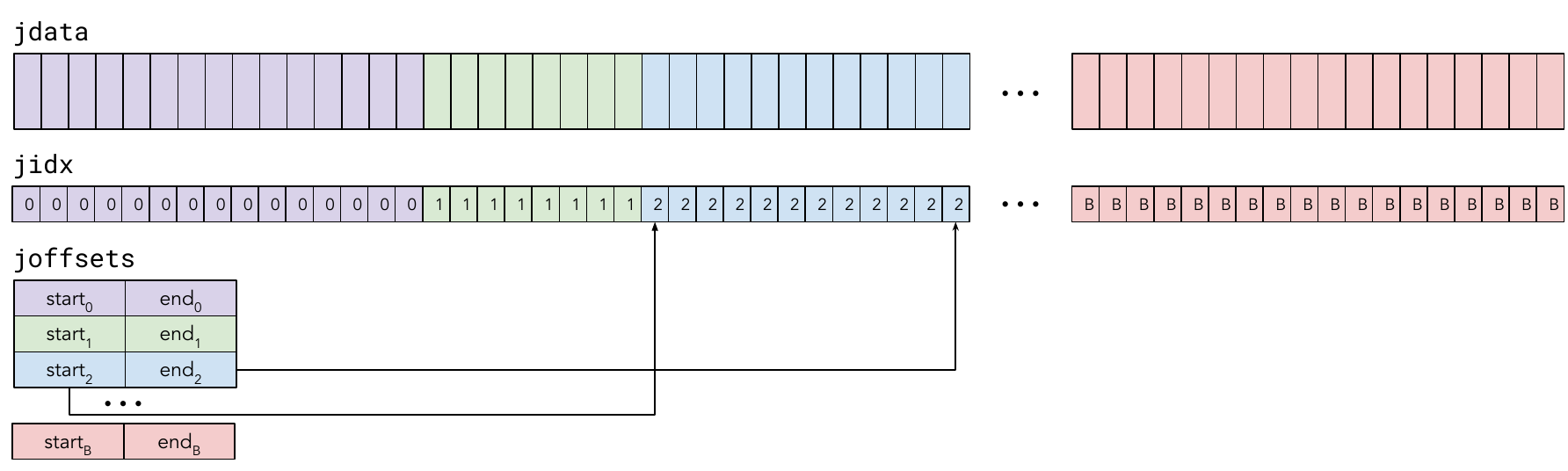}
    \caption{Conceptually, a \JagT is a list of tensors whose first dimension differs between each list item and all other dimensions match. Internally, \JagT concatenates all the tensors in the list into one dense tensor (\jdata) and stores two auxiliary pieces of metadata: \jidx, which specifies the index in the list that each tensor is in and \joffsets, which specifies the start and end indexes of each list item in \jdata.}
    \label{fig:jaggedTensorLayout}
\end{figure*}

In general, we cannot expect each grid within a minibatch to have the same number of voxels. Thus, \longnamespaced must expose operations on \emph{jagged} arrays of data. \longnamespaced exposes the \JagT class for this purpose. 
Conceptually a \JagT can be thought of a list of tensors $[t_1, t_2, \ldots t_B]$ where each tensor $t_i$ has shape $[N_i, *]$ \ie each tensor has different first dimension but matches in subsequent dimensions. For example, if a \JagT represents per-voxel attributes in a batch of grids, then $N_i$ will be the number of voxels in the $i^\text{th}$ grid in the batch. Under the hood, \longnamespaced efficiently encodes these tensors contiguously in memory to enable fast operators on them. Specifically, a \JagT consits of three parts: 

\begin{enumerate}[label=\alph*)]
    \item \jdata which is a $[N_1 + \ldots + N_B, *]$-shaped tensor equivalent to concatenating $t_1, \ldots t_B$ along their first axis
    \item \joffsets which is a $[B, 2]$-shaped tensor such that \\ $\text{\texttt{joffsets}}[i, :]$ is the start and end tensor $t_i$ in \jdata
    \item \jidx which is a $[N_1 + \ldots N_B]$-shaped tensor such that \texttt{jidx}[i] is the index (from $0$ to $B-1$) of the $i^\text{th}$ element in \jdata

\end{enumerate}
Figure~\ref{fig:jaggedTensorLayout} shows this layout pictorially. Note that \joffsets and \jidx are also available for \GridB since these represent a jagged collection of voxels. In the subsequent paragraphs, a tensor shape of $-1$ refers to a jagged dimension. For example, a \JagT containing the voxel coordinates of a \GridB would have shape $[B, -1, 3]$.

\subsubsection{List of Operators}
\longnamespaced supports a range of differentiable operators on minibatches of sparse voxel grids of tensor data. These operators are written in CUDA and C++ and interoperate with PyTorch. Here we give a high-level description of the major operators in \longname. A concise summary of these are given in Table~\ref{tab:features}.

\paragraph{Grid Construction Operators} A \GridB in \longnamespaced can be created from a \JagT of point clouds; voxel ($ijk$) coordinates; triangle meshes (the set of voxels which intersect a mesh); other \GridBs via padding, coarsening, or subdivision; and from dense grids with masks. 

\paragraph{Sampling Operators} A common operator is to sample tensor values on a voxel grid at a set of query points $Q \in 2^{\mathbb{R}^3}$. \longnamespaced provides differentiable sampling operators which accept a \GridB $G$, a \JagT of per-voxel features $Z$ with shape $[B, -1, *]$, and a \JagT of query points $Q$ with shape $[B, -1, 3]$. These operators return a set of features $Z_Q$ sampled at each point $q \in Q$ using Trilinear or B\'ezier interpolation.

\paragraph{Splatting Operators} \longnamespaced supports splatting data stored at points onto a grid using Trilinear or B\'ezier interpolation. These operators accept a \GridB $G$, a \JagT $P$ of points, and a \JagT $Z$ of per-point features. They produce a \JagT of features (one per voxel in $G$) by splatting the feature at each point onto the neighboring voxels.

\paragraph{Convolution, Pooling, Upsampling, and Attention} \longnamespaced supports sparse convolution via a novel accelerated implementation (Section~\ref{sec:sparse-conv}). The convolution operator accepts a \GridB $G_\text{in}$, a kernel $K$, and a \JagT of features $Z_\text{in}$ and produces a \GridB $G_\text{out}$, and \JagT $Z_\text{out}$ by performing sparse convolution. We further support average and max pooling operators on a \GridB and \JagT pair as well as an upsampling operator which upsamples a \GridB and \JagT of features via subdivision and nearest neighbor sampling. \longnamespaced supports attention by calling out to Flash Attention~\cite{dao2022flashattention} on a \JagT.

\paragraph{Ray Marching} \longnamespaced comes with a number of operators for intersecting rays with grids. These include enumerating the set of voxels along a ray, parameterized by intervals of $t$ along a ray which intersect a grid; finding the intersection between rays and the level set of an implicit function stored on a grid; and volume rendering. Ray marching operations are implemented using a hierarchical DDA algorithm outlined in Section~\ref{sec:hdda}.

\newcommand*\rot{\rotatebox{90}}

\begin{table}[!ht]
\caption{Feature comparison between our \longnamespaced and four alternative sparse DL frameworks that represent state-of-the-art.\label{tab:features}}
\begin{adjustbox}{width=\columnwidth}
    \begin{threeparttable}
    \centering
    
    \rowcolors{2}{}{gray!10}\
    \scriptsize
    \begin{tabular}{lcccccc}\toprule

    &\rot{\textbf{\longname}} &\rot{Torchsparse++}\rot{~\cite{tang2022torchsparse}} &\rot{Minkowski Engine}\rot{~\cite{choy20194d}} &\rot{SpConv}\rot{~\cite{spconv2022}} &\rot{NerfAcc}\rot{~\cite{li2023nerfacc}} \\\midrule
    \cellcolor[HTML]{A8A8A8}\textbf{Grid Construction Methodology} &\cellcolor[HTML]{A8A8A8} &\cellcolor[HTML]{A8A8A8} &\cellcolor[HTML]{A8A8A8} &\cellcolor[HTML]{A8A8A8} &\cellcolor[HTML]{A8A8A8} \\
    Coordinate Lists & \CHECK & \CHECK & \CHECK & \CHECK & - \\
    Dense Grids & \CHECK & - & \CHECK & \CHECK & \CHECK \\
    Pointclouds & \CHECK & \CHECK\tnote{1} & \CHECK & \CHECK\tnote{1} & - \\
    Meshes & \CHECK &- &- &- &- \\
    Dual Grid & \CHECK &- &- &- &- \\
    \cellcolor[HTML]{A8A8A8}\textbf{Grid Topology Feature Set} &\cellcolor[HTML]{A8A8A8} &\cellcolor[HTML]{A8A8A8} &\cellcolor[HTML]{A8A8A8} &\cellcolor[HTML]{A8A8A8} &\cellcolor[HTML]{A8A8A8} \\
    Spatial Dimensions &3 &3 &Arbitrary &3 &3 \\
    Subdivision + Coarsening & \CHECK & \CHECK & \CHECK & \CHECK & - \\
    Adaptive Subdivision & \CHECK &- &- &- &- \\
    Device Accelerated Grid Building & \CHECK &- &- &- &- \\
    Mutable Grids & \CHECK &- &- &- &- \\
    Zero-Copy Grid Cropping & \CHECK &- &- &- &- \\
    \cellcolor[HTML]{A8A8A8}\textbf{Indexing and Sampling} &\cellcolor[HTML]{A8A8A8} &\cellcolor[HTML]{A8A8A8} &\cellcolor[HTML]{A8A8A8} &\cellcolor[HTML]{A8A8A8} &\cellcolor[HTML]{A8A8A8} \\
    $\text{Point}\rightarrow\text{Grid Sampling}$ & \CHECK & - & \CHECK & - & - \\
    \hspace{3mm}Trilinear Interpolation & \CHECK & - & \CHECK & - & - \\
    \hspace{3mm}Bézier Interpolation & \CHECK &- &- &- &- \\
    \hspace{3mm}Gradient Sampling Support & \CHECK & - & \CHECK & - & - \\
    $\text{Grid}\rightarrow\text{Point Splatting}$ & \CHECK &- &- &- &- \\
    \hspace{3mm}Trilinear Interpolation & \CHECK &- &- &- &- \\
    \hspace{3mm}Bézier Interpolation & \CHECK &- &- &- &- \\
    Accelerated Spatial Neighbour Indexing & \CHECK &- &- &- &- \\
    \cellcolor[HTML]{A8A8A8}\textbf{Geometry Functionality} &\cellcolor[HTML]{A8A8A8} &\cellcolor[HTML]{A8A8A8} &\cellcolor[HTML]{A8A8A8} &\cellcolor[HTML]{A8A8A8} &\cellcolor[HTML]{A8A8A8} \\
    Point/Voxel Intersections & \CHECK &- &- &- &- \\
    Cube/Voxel Intersections & \CHECK &- &- &- &- \\
    Marching Cubes Mesher & \CHECK &- &- &- &- \\
    \cellcolor[HTML]{A8A8A8}\textbf{Raytracing Feature Set} &\cellcolor[HTML]{A8A8A8} &\cellcolor[HTML]{A8A8A8} &\cellcolor[HTML]{A8A8A8} &\cellcolor[HTML]{A8A8A8} &\cellcolor[HTML]{A8A8A8} \\
    Ray Sampling & \CHECK &- &- &- & \CHECK \\
    Implicit Field Intersection & \CHECK &- &- &- & \CHECK \\
    HDDA Device-Accelerated Raytracing & \CHECK &- &- &- &- \\
    \cellcolor[HTML]{A8A8A8}\textbf{ML Operators} &\cellcolor[HTML]{A8A8A8} &\cellcolor[HTML]{A8A8A8} &\cellcolor[HTML]{A8A8A8} &\cellcolor[HTML]{A8A8A8} &\cellcolor[HTML]{A8A8A8} \\
    Sparse Convolution & \CHECK & \CHECK & \CHECK & \CHECK & - \\
    Pooling & \CHECK &- & \CHECK & \CHECK & - \\
    Flash Attention & \CHECK &- &- &- &- \\
    Gaussian Splatting & \CHECK &- &- &- &- \\
    \cellcolor[HTML]{A8A8A8}\textbf{Interoperability} &\cellcolor[HTML]{A8A8A8} &\cellcolor[HTML]{A8A8A8} &\cellcolor[HTML]{A8A8A8} &\cellcolor[HTML]{A8A8A8} &\cellcolor[HTML]{A8A8A8} \\
    PyTorch Extension & \CHECK & \CHECK & \CHECK & \CHECK & \CHECK \\
    Interoperable Volume Format & \CHECK &- &- &- &- \\
    Modeling/Manipulation Toolset & \CHECK &- &- &- &- \\
    Shared Datamodel w/DCC Applications & \CHECK &- &- &- &- \\
    Shared Datamodel w/Industry Renderers & \CHECK &- &- &- &- \\
    \bottomrule

    \end{tabular}

    \begin{tablenotes}\footnotesize
    \footnotesize \item[1] Supported via quantization of points and index-based construction
    \end{tablenotes}

    \end{threeparttable}
\end{adjustbox}
\end{table}

\section{Experiments}
In this section, we demonstrate the effectiveness of \longnamespaced through a series of benchmarks and qualitative examples of use cases. Our experiments demonstrate that our framework successfully covers a broad variety of use cases and operations, while achieving state-of-the-art runtime performance and memory efficiency. First, we perform micro-benchmarks of the most important operators in \longname, comparing them against corresponding state-of-the-art operators in other sparse deep learning frameworks in terms of both memory usage and speed. Next, we run a macro-benchmark showing that \longnamespaced remains performant in the real-world use case of training a sparse convolutional neural network (CNN). Finally, we demonstrate the utility of \longnamespaced by showing its use in several key applications on high-resolution 3D data. These applications include  3D reconstruction from points, semantic completion, 3D shape generation, and neural radiance field rendering.

\subsection{Micro-benchmarks}
We evaluate the runtime performance and memory efficiency of the core primitive operations in \longname, comparing against operators available in other frameworks. First, we compare the speed and memory footprint of our core algorithm for index grid construction, which converts a list of \verb|ijk| integer or \verb|xyz| point coordinates to a VDB IndexGrid on the GPU. All grid construction operations (\eg from meshes) make use of this build algorithm, so this is a crucial benchmark. Second, we evaluate the performance of our HDDA ray marching algorithm, which is the backbone of all ray-tracing algorithms in the framework. Finally, we evaluate the performance of our convolution operator on a novel benchmark consisting of a variety of real-world examples spanning different sparsity patterns and channel depths.

Each data-point for the experiments on grid construction and convolution, sections \ref{sec:exp:gridconstruction} and \ref{sec:exp:conv} respectively, were averaged from the 4 best runs out of 5 runs to mitigate outliers.  Between each run we made sure to clear the device's L2 cache to make sure that no framework was benefiting from the uneven advantages of a warm cache.  The experiment in sections \ref{sec:exp:conv} was run on a machine with an AMD 7950X 16-Core CPU and GeForce RTX 4090 GPU, with 128GB of host memory and 24GB of device memory.  The experiment in section \ref{sec:exp:gridconstruction} was run on a machine with an AMD 3975WX 32-Core CPU and RTX 6000 Ada Generation GPU, with 128GB of host memory and 48 GB of device memory.

The experiment on ray marching in section \ref{sec:exp:HDDA} was performed by averaging the results of 1,000 runs where each run consisted of casting 1,024 rays.  This experiment was run on a machine with an AMD 3975WX 32-Core CPU and GeForce RTX 3090 Ti GPU, with 128GB of host memory and 24GB of device memory.

\subsubsection{IndexGrid Construction}
\label{sec:exp:gridconstruction}
\begin{figure}
    \centering
    \includegraphics[width=0.99\linewidth]{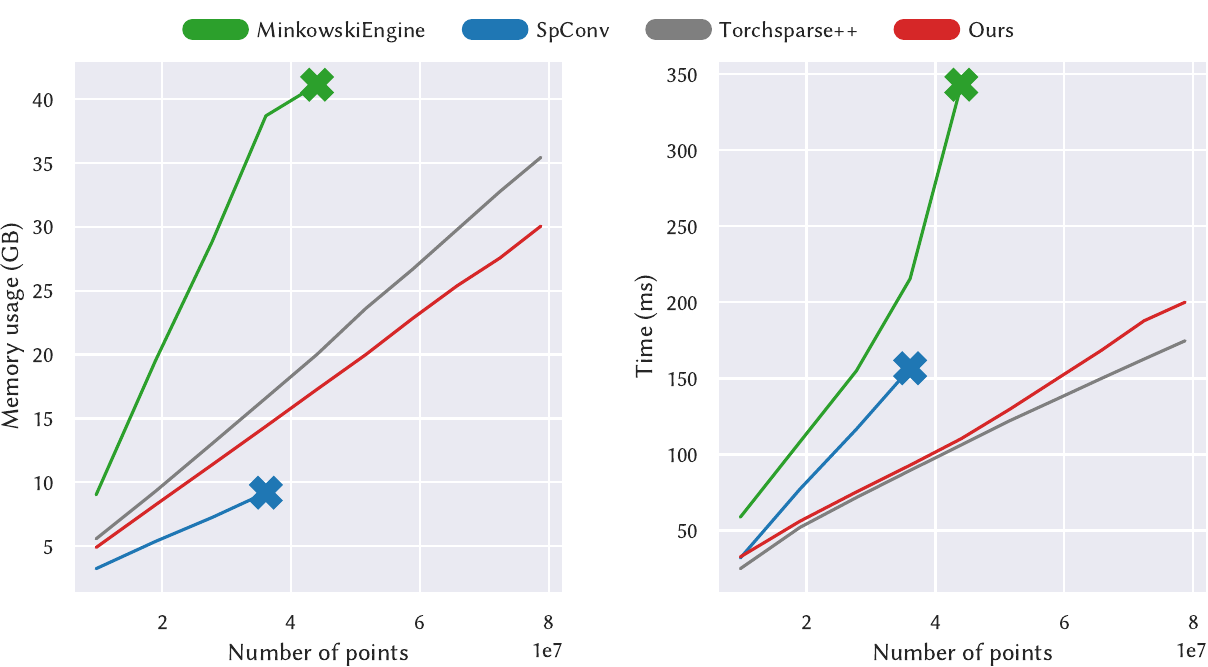}
    \caption{\textbf{Left:} Maximum memory utilization (y-axis) when constructing a grid for a given number of coordinates (x-axis). While MinkowskiEngine and spconv fail as the problem size grows larger, ours (\longnamespaced) remains more memory efficient relative to TorchSparse++. \textbf{Right:} Runtime (y-axis) required to construct a grid for a given number of coordinates (x-axis). Our method is on par with TorchSparse++ and significantly faster than MinkowskiEngine and spconv especially when scaling to much larger inputs. The cross marks on MinkowskiEngine indicate an out of memory error, while those on spconv, where we otherwise seem to be within memory limits, indicate an illegal memory access exception that occurs within the framework at these point counts. }
    \label{fig:micro-build}
\end{figure}

The IndexGrid construction algorithm, detailed in Section~\ref{sec::GridBuild}, converts a list of \verb|ijk| integer or \verb|xyz| point coordinates into a VDB IndexGrid on the GPU. It forms the backbone of all grid constructions in \longname, while also acting as a means to initialize sparse grids. We evaluate the runtime performance and memory footprint of our grid construction algorithm against the analogous calculations in TorchSparse++~\cite{tangandyang2023torchsparse}, MinkowskiEngine~\cite{choy20194d}, and spconv~\cite{spconv2022} by constructing a grid with random points sampled from a normal distribution followed by any auxiliary computation required to execute the sparse convolution. Figure~\ref{fig:micro-build} shows the maximum memory usage and runtime of this process as a function of the number of input points. Our method runs significantly faster than MinkowskiEngine and spcov and on par with TorchSparse++. Moreover, our approach is more memory efficient than TorchSparse++ while MinkowskiEngine and spconv both fail as the problem size increases. Thus, \longnamespaced can process input data much larger than current state-of-the-art sparse DL frameworks. We conducted this experiment on an NVIDIA A100 GPU.

\subsubsection{Hierarchical DDA}
\label{sec:exp:HDDA}
We profile our HDDA ray marching on a 3-voxel-wide narrow-band level set of the Stanford bunny extracted at various (effective) resolutions ranging from $32^3$ to $1024^3$. The ray marching axis-aligned bounding box of the bunny is 1.2x of its tight axis-aligned bounding box and all rays are always marched through the entire volume constructing intervals along the ray. We compare our algorithm with the widely used NerfAcc~\cite{li2023nerfacc} library (e.g. by NeRFStudio~\cite{nerfstudio}) for ray marching and volume rendering. NerfAcc provides a highly optimized DDA over a dense binary grid implemented in CUDA. Table~\ref{tab:hdda} shows that \longnamespaced constantly achieves 1.5x to 3x faster runtimes than NerfAcc while maintaining a comparable or lower (up to 100x at high resolutions) memory footprint. The same conclusion applies to the real-world scene as well, where in the large-scale NeRF application (\S\ref{sec:nerf}) we observe 1.3x faster ray marching with \longnamespaced comparing to NerfAcc, and 30x less memory footprint at effective $1024^3$ resolution on the Laguna Seca Raceway scene.

\begin{table}[!t]

\centering

\caption{Comparison between HDDA ray marching in \longnamespaced and DDA ray marching in NerfAcc~\cite{li2023nerfacc} on the 3-voxel-wide shell of the Stanford bunny. Our approach consistently outperforms that of NerfAcc by 1.5x to 3x on runtime while also maintaining up to 100x lower GPU memory footprint.}
\label{tab:hdda}

\begin{adjustbox}{width=\columnwidth}

\begin{tabular}{lcccccc}
    \toprule
    Grid Resolution & $32^3$ & $64^3$ & $128^3$ & $256^3$ & $512^3$ & $1024^3$ \\
    \midrule
    \textbf{\underline{Rays / Sec (M)}} & & & & & & \\
    NerfAcc & 2.57 & 2.46 & 2.09 & 1.41 & 0.82 & 0.47 \\
    \longnamespaced & 3.77 & 3.40 & 2.81 & 2.24 & 1.83 & 1.43 \\[0.3em]
    \textbf{\underline{GPU Mem. (MB)}} & & & & & & \\
    NerfAcc & 0.24 & 0.41 & 2.15 & 16.3 & 129 & 1028 \\
    \longnamespaced & 0.38 & 0.37 & 0.40 & 0.79 & 2.46 & 8.85 \\[0.3em]
     \textbf{\underline{Cell Intersections/Ray}}& 0.71 & 0.72 & 0.73 & 0.78 & 0.70 & 0.66 \\
    \bottomrule
\end{tabular}
\end{adjustbox}
\end{table}

\subsubsection{Sparse Convolution}
\label{sec:exp:conv}
\begin{figure}
    \centering
    \includegraphics[width=0.99\linewidth]{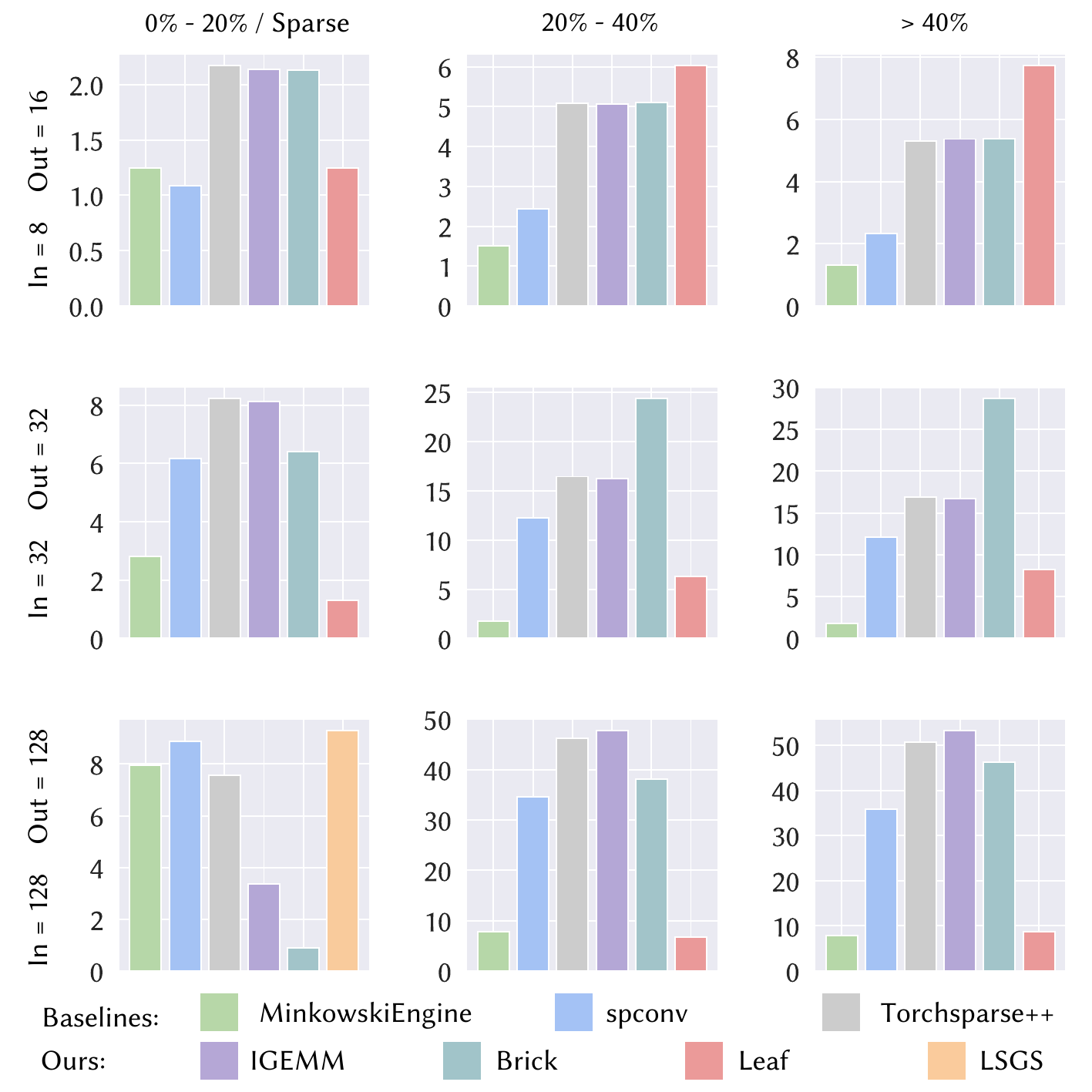}
    \caption{Micro-benchmarks of our convolution operator. The three columns represent increasing leaf node-level occupancy reflecting grid sparsity, while the three rows denote different input (In) and output (Out) channel sizes. The bars represent speed measured in "effective" TFLOPs [Section \ref{sec:exp:conv}]. To incorporate extreme sparse cases we use the KITTI dataset for the lower-left subplot.}
    \label{fig:micro-conv}
\end{figure}

We profile our core convolution operators across a range of different feature depths: A low-depth regime with input depth of 8 and output depth of 16, a medium depth case with input and output depths of 32, and a high-depth scenario with input and output depths of 128. Orthogonal to feature depth, we examine three different degrees of sparsity:
\begin{enumerate}[label=\alph*)]
\item a highly sparse regime leading to voxel occupancy (at the IndexGrid leaf node level) below 20\%, harvested from typical single-scan LiDAR datasets of rasterized point clouds \cite{behley2021ijrr}
\item a case of moderate leaf node-level occupancy of 20-40\%, originating from rasterized surfaces, and
\item a case of higher density stemming from rasterization of volumetric data with nontrivial codimensional thickness, with leaf node-level occupancy in excess of 40\%
\end{enumerate}

The performance plots in Figure \ref{fig:micro-conv} include four implementations available in our framework:
\begin{enumerate}[label=\alph*)]
\item an adaptation of SpConv v2 (labeled \emph{IGEMM}) that employs our tree-derived indexing scheme instead of a spatial hash
\item local densification at the leaf-node level (Scenario 1 in Section \ref{sec:sparse-conv}; labeled \emph{Leaf} in the figure)
\item local densification at a $4\times 2\times 2$ ``brick'' (Scenario 2 in Section \ref{sec:sparse-conv}; labeled \emph{Brick} in the figure)
\item the shared-memory Local Gather-GEMM-Scatter paradigm of scenario 3 in Section \ref{sec:sparse-conv} (labeled \emph{LGGS} in the figure); this last option is only leveraged for high-depth convolution operations
\end{enumerate}
As can be surmised from Figure \ref{fig:micro-conv}, these four approaches allow us to select an operator implementation that is the most competitive to alternatives (i.e. those not incorporated as possible backends in \longname) in each case.
We note that in our experiments, optimizations beyond the \emph{IGEMM} baseline were deployed when appropriate as part of the inference pipeline only; for training we defaulted to the \emph{IGEMM} option for simplicity and as to avoid further specialization of the gradient computation for the filter coefficients.

Our benchmark also indicates the TFLOPS achieved by the top performer in each instance. This is an ``effective'' TFLOPs figure that reflects the method's degree of success in leveraging both spatial sparsity, and stencil sparsity (e.g. avoiding, to the degree possible, unnecessary multiply-and-accumulate (MAC) operations for stencil weights that are absent at specific grid locations). We compute this ``effective TFLOPS'' figure by counting the bare minimum number of operations essential for the stencil application, excluding from this count operations that would be associated with null weights. These numbers should be contrasted with the architectural ceiling of 73TFLOPS (or 82.6TFLOPS with a boost clock) on the RTX 4090 platform used in these experiments.

\subsection{Macro-benchmarks}
\label{sec:exp:macro}

\subsubsection{Full Network Inference}
\begin{figure*}
    \centering
    \includegraphics[width=0.99\linewidth]{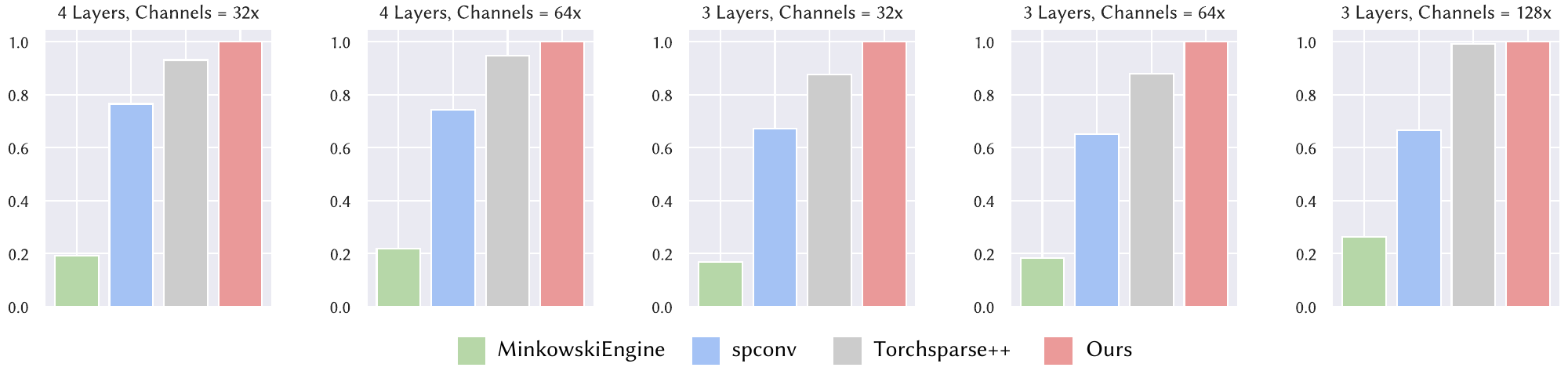}
    \caption{End-to-end speed comparison of \longnamespaced with state-of-the-art sparse frameworks under different configurations of the XCube~\cite{ren2023xcube} backbone. Runtime speed is normalized over the best model (always ours in this case), and the higher the better. Results are averaged over 10 runs.}
    \label{fig:macro-xcube}
\end{figure*}

We benchmark the end-to-end performance of \longnamespaced-based network inference.
To this end, we leverage the generative backbone of XCube from \cite{ren2023xcube}.
Such a backbone has a typical encoder-decoder structure and is representative for sparse U-Net designs by first applying a set of downsampling operations to reduce spatial resolution and then upsampling to the original scale.
Our dataset is based on a voxelized version of the KartonCity~\cite{kartoncity} dataset containing 500 representative samples, where we uniformly pick spatial resolutions from 256, 512, and 1024. This dataset contains dense geometry of a synthetic city that is suitable for generative tasks. Detailed speed comparison on different configurations of the network are shown in Figure \ref{fig:macro-xcube}. We consistently perform better than the state-of-the-art baselines under different spatial resolutions and channel sizes.
Our results were averaged from the 4 best runs out of 5 runs to mitigate outliers.  Between each run we made sure to clear the device's L2 cache to make sure that no framework was benefiting from the uneven advantages of a warm cache.  The experiment was run on a machine with an AMD 7950X 16-Core CPU and GeForce RTX 4090 GPU, with 128GB of host memory and 24GB of device memory.

\subsubsection{Neural Radiance Fields}
We run the full end-to-end neural radiance fields training and testing session based on a reference implementation of Instant-NGP (iNGP)~\cite{muller2022instant}. In order to query the color of a sampled ray, one would first perform ray marching through the scene to obtain samples close to the scene surface. The features at the sample positions are then retrieved and volume rendered to aggregate the final color. In \cite{muller2022instant}, a cascade of binary grids of varying voxel sizes is used to represent the rough sparsity of the scene. By replacing the cascaded grid structure with the \longnamespaced grid representation, we can accelerate the process of ray marching using the HDDA algorithm as introduced, while benefiting from the modest memory consumption provided by the VDB data structure. We run the neural radiance fields on a GeForce RTX 4090 GPU on one scene in the Waymo Open Dataset \cite{waymo}. The training speed of ours compared to iNGP is 26.1it/s vs. 26.4it/s, while the inference speed of ours compared to iNGP is 1.90FPS vs 1.62FPS. As \longnamespaced is initialized from LiDAR point clouds and offers more precise locations of the samples, we reached a test PSNR of 27.07, in comparison to 25.89 for iNGP.

\subsection{Example Applications}\label{sec:applications}
We demonstrate that \longnamespaced is a practical tool for building real-world 3D deep learning applications. Here we present several applications of \longnamespaced, some of which are reimplementations of published works. These include large scale surface reconstruction from point clouds using NKSR~\cite{huang2023nksr}, high resolution hierarchical object and scene generation using XCube~\cite{ren2023xcube}, large-scale Neural Radiance Fields, and Deep-Learning based simulation super-resolution.

\subsubsection{Large-scale Surface Reconstruction}
\begin{figure*}
    \centering
    \includegraphics[width=\linewidth]{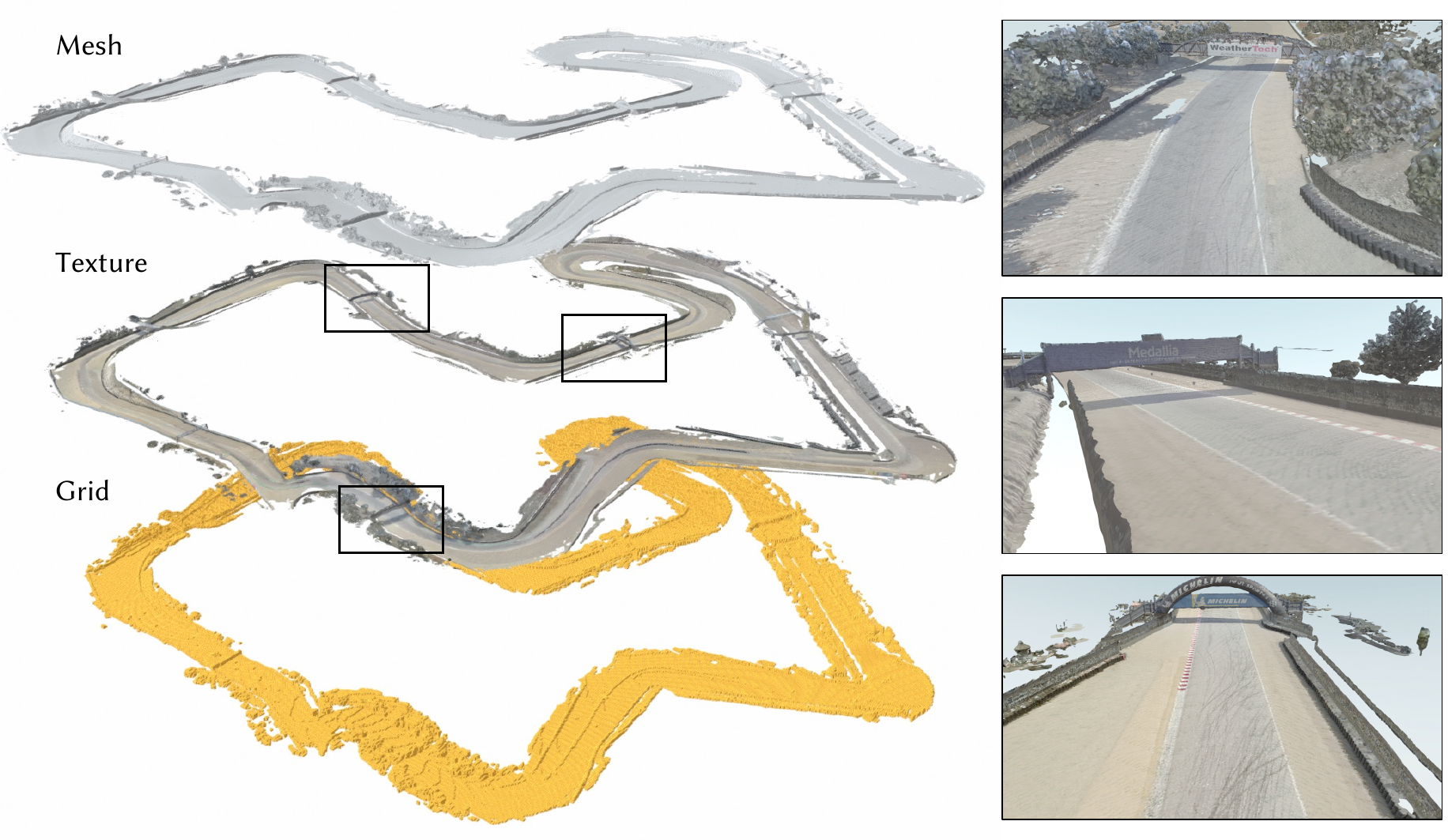}
    \caption{\longnamespaced helps state-of-the-art surface reconstruction models to scale to larger scenes spanning kilometers. Textures can be re-projected from images in a faster way with the help of our rendering operators.}
    \label{fig:nksr}
\end{figure*}

NKSR~\cite{huang2023nksr} uses a sparse voxel hierarchy to encode a neural field of features which are used to perform a learned kernel ridge regression to solve a variational surface reconstruction problem from oriented point clouds. NKSR achieves state-of-the-art reconstruction and generalization results. We fully re-implemented NKSR using \longnamespaced replacing the convnet with our implementation, the meshing with our marching cubes implementation, and implementing a batched Kernel Ridge Regression solver as an \longnamespaced C++ extension. We remark that this extension is a single file consisting of a few hundred lines of code which only depends on PyTorch and \longname. Figure~\ref{fig:nksr} shows a mesh reconstructed using our implementation from 350 million input points. This reconstruction took 2 minutes on 8 V100 GPUs.

\subsubsection{3D Generative Models}
\begin{figure*}
    \centering
    \includegraphics[width=0.99\linewidth]{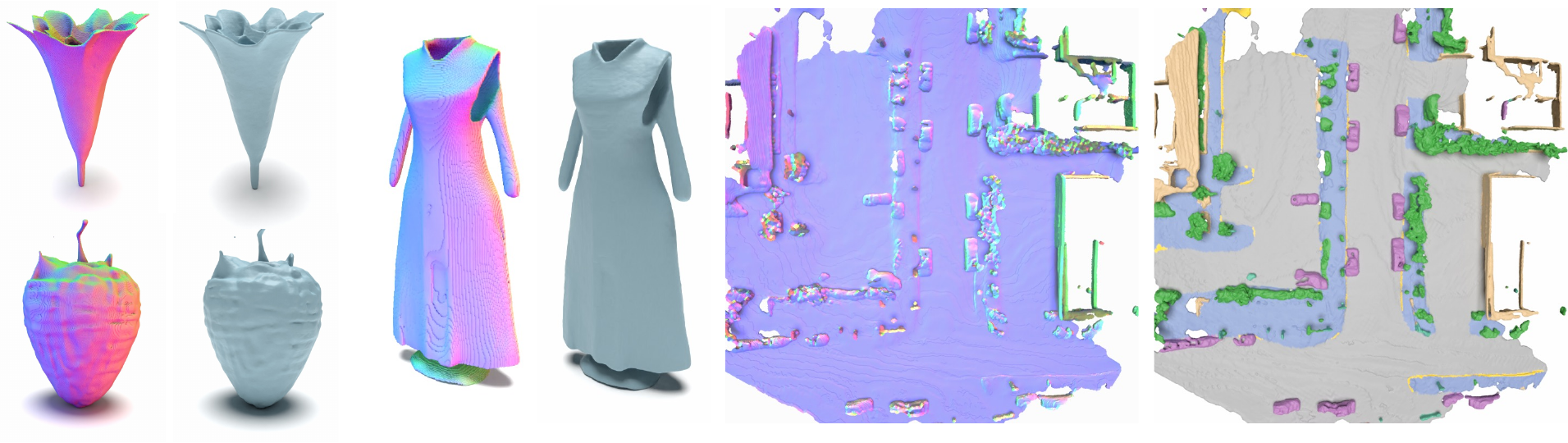}
    \caption{\longnamespaced helps push the limit of 3D generative models in terms of resolution and scale. With less memory usage and faster speed, we generate high-resolution 3D objects ($512^3$) and large-scale scenes ($1024^3$). We provide the generated sparse voxel grid colored by normal and the extracted mesh for each sample.}
    \label{fig:xcube}
\end{figure*}

\looseness=-1 We used \longnamespaced to re-implement XCube \cite{ren2023xcube}, a 3D generative model for high-resolution voxel hierarchies of objects and scenes. XCube benefits directly when using \longnamespaced to enable it to train on datasets with substantially larger footprints and higher spatial resolution while consuming less GPU memory. With the support of \longname, XCube can be scaled up to spatial scale of 100m $\times$ 100m at 10cm resolution. Figure~\ref{fig:xcube} demonstrates unconditional generation of high-resolution 3D objects trained using the Objaverse~\cite{Objaverse} dataset and large-scale outdoor scenes trained on the Waymo~\cite{waymo} dataset.

\subsubsection{Large-scale Neural Radiance Fields}
\label{sec:nerf}

\begin{figure*}
    \centering
    \includegraphics[width=0.99\linewidth]{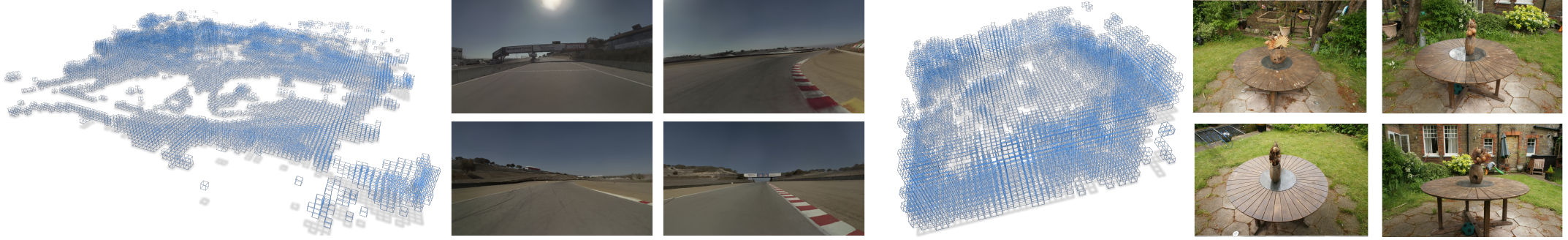}
    \caption{\longnamespaced can be used to support large-scale Neural Radiance Fields (NeRF) training and rendering by providing a memory efficient acceleration structure for spatial skipped ray marching.
    Here we showcase the \longnamespaced grid and NeRF renderings on two scenes, where the left one is our capture of the
    Laguna Seca Raceway (1km squared area) and the right one is the Garden scene from the Mip-NeRF 360 dataset~\cite{barron2022mip360}.
    }
    \label{fig:nerf}
\end{figure*}

\longnamespaced can be used to support large-scale Neural Radiance Fields by providing a memory efficient acceleration structure for spatial skipped ray marching. Figure~\ref{fig:nerf} provides two showcases of this application including a 1km squared area capture of the Laguna Seca Raceway and a standard Garden scene in the NeRF literature from Mip-NeRF 360 dataset~\cite{barron2022mip360}.

\subsubsection{Simulation Super-Resolution}
\begin{figure*}
    \centering
    \includegraphics[width=0.99\linewidth]{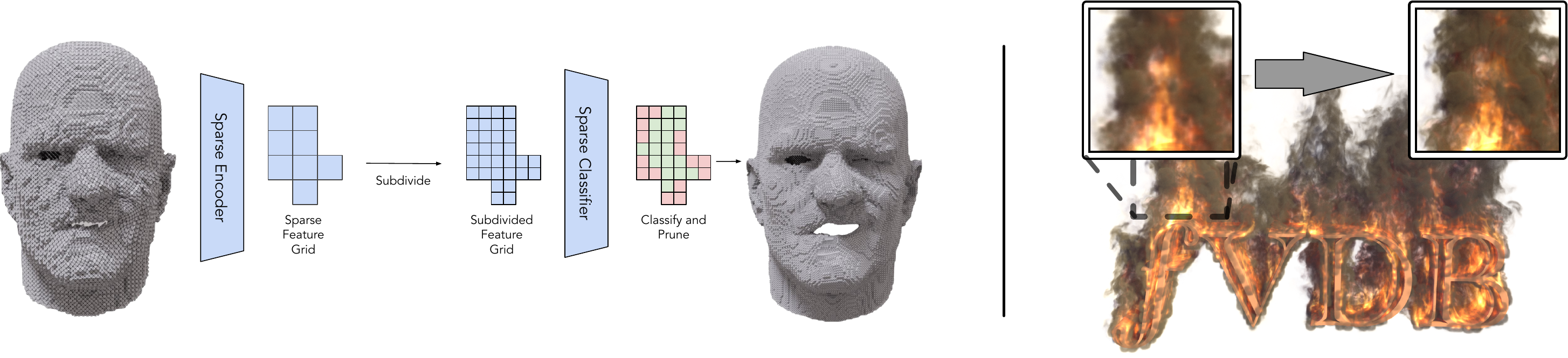}
    \caption{\longnamespaced can enable super-resolution applications on unbounded, sparse physical simulation data where previous methods using traditional dense data structures and operations were prohibitive on large-scale, sparse simulation data.  Here we showcase some of the methods we are applying to simulations of fire and facial muscle + skin simulations to super-resolve details using fully convolutional network architectures utilizing \longnamespaced operators.  }
    \label{fig:sr_face}
\end{figure*}

\longnamespaced can enable novel applications of super-resolution techniques to inherently sparse, 3D data such as those produced by physical simulations which operate in unbounded domains. Previous approaches can be memory constrained and computationally prohibitive for large domains if approached with dense data structures and operators.  Figure~\ref{fig:sr_face} shows preliminary results of work we are currently undertaking which trains fully convolutional super-resolution networks such as DCSRN~\cite{chen_brain_2018} and 3D-FSRCNN~\cite{mane_image_2020} with operators implemented in \longname.  Currently in development are super-resolution models for several simulation domains including muscle and skin dynamics as well as fluid simulations.

\section{Conclusion and Future Work}
We presented \longname, a novel GPU-optimized framework for deep learning on large-scale 3D data. Our framework includes a broad set of novel differential primitives which can be used to build deep-learning pipelines for a wide variety of 3D tasks. These primitives include GPU accelerated grid building, ray marching, convolution, sampling, splatting, etc. Furthermore, \longnamespaced has a significantly more comprehensive suite of features than existing frameworks, runtime performance that is at-par or superior to state-of-the-art and memory efficiency that exceeds state-of-the-art by a large margin. \longnamespaced uses a single, novel VDB IndexGrid data structure to accelerate all operations, making it composable and easily extensible. We demonstrated the effectiveness of \longnamespaced via extensive quantitative benchmarks and qualitative demonstrations on real-world 3D learning use cases, showing that \longnamespaced enables high-performance deep learning on large scale 3D data. 

In the future, we plan to extend \longnamespaced with more differentiable operators such as hierarchical dual marching cubes, and particle/blob to grid conversion functions (for differentiable physics and particle rendering \eg Gaussian Splatting~\cite{kerbl3Dgaussians}). We further plan to develop a high level utility library of neural network architectures for common tasks that can be used off-the-shelf for downstream applications. Beyond new features, an exciting avenue of future work which can lead to even greater sparse convolution performance is to dispatch the optimal kernel on a per-leaf basis depending on local sparsity pattern. Finally, we plan to release the code for \longnamespaced as open-source software expeditiously following publication.

\bibliographystyle{ACM-Reference-Format}
\bibliography{bibliography}


\begin{thebibliography}{41}


\ifx \showCODEN    \undefined \def \showCODEN     #1{\unskip}     \fi
\ifx \showDOI      \undefined \def \showDOI       #1{#1}\fi
\ifx \showISBNx    \undefined \def \showISBNx     #1{\unskip}     \fi
\ifx \showISBNxiii \undefined \def \showISBNxiii  #1{\unskip}     \fi
\ifx \showISSN     \undefined \def \showISSN      #1{\unskip}     \fi
\ifx \showLCCN     \undefined \def \showLCCN      #1{\unskip}     \fi
\ifx \shownote     \undefined \def \shownote      #1{#1}          \fi
\ifx \showarticletitle \undefined \def \showarticletitle #1{#1}   \fi
\ifx \showURL      \undefined \def \showURL       {\relax}        \fi
\providecommand\bibfield[2]{#2}
\providecommand\bibinfo[2]{#2}
\providecommand\natexlab[1]{#1}
\providecommand\showeprint[2][]{arXiv:#2}

\bibitem[kar(2023)]%
        {kartoncity}
 \bibinfo{year}{2023}\natexlab{}.
\newblock \bibinfo{title}{3D Karton City model}.
\newblock
  \bibinfo{howpublished}{\url{https://www.turbosquid.com/3d-models/3d-karton-city-2-model-1196110}}.
\newblock
\newblock
\shownote{Accessed: 2023-08-01}.


\bibitem[Abadi et~al\mbox{.}(2015)]%
        {tensorflow2015-whitepaper}
\bibfield{author}{\bibinfo{person}{Mart\'{i}n Abadi}, \bibinfo{person}{Ashish
  Agarwal}, \bibinfo{person}{Paul Barham}, \bibinfo{person}{Eugene Brevdo},
  \bibinfo{person}{Zhifeng Chen}, \bibinfo{person}{Craig Citro},
  \bibinfo{person}{Greg~S. Corrado}, \bibinfo{person}{Andy Davis},
  \bibinfo{person}{Jeffrey Dean}, \bibinfo{person}{Matthieu Devin},
  \bibinfo{person}{Sanjay Ghemawat}, \bibinfo{person}{Ian Goodfellow},
  \bibinfo{person}{Andrew Harp}, \bibinfo{person}{Geoffrey Irving},
  \bibinfo{person}{Michael Isard}, \bibinfo{person}{Yangqing Jia},
  \bibinfo{person}{Rafal Jozefowicz}, \bibinfo{person}{Lukasz Kaiser},
  \bibinfo{person}{Manjunath Kudlur}, \bibinfo{person}{Josh Levenberg},
  \bibinfo{person}{Dandelion Man\'{e}}, \bibinfo{person}{Rajat Monga},
  \bibinfo{person}{Sherry Moore}, \bibinfo{person}{Derek Murray},
  \bibinfo{person}{Chris Olah}, \bibinfo{person}{Mike Schuster},
  \bibinfo{person}{Jonathon Shlens}, \bibinfo{person}{Benoit Steiner},
  \bibinfo{person}{Ilya Sutskever}, \bibinfo{person}{Kunal Talwar},
  \bibinfo{person}{Paul Tucker}, \bibinfo{person}{Vincent Vanhoucke},
  \bibinfo{person}{Vijay Vasudevan}, \bibinfo{person}{Fernanda Vi\'{e}gas},
  \bibinfo{person}{Oriol Vinyals}, \bibinfo{person}{Pete Warden},
  \bibinfo{person}{Martin Wattenberg}, \bibinfo{person}{Martin Wicke},
  \bibinfo{person}{Yuan Yu}, {and} \bibinfo{person}{Xiaoqiang Zheng}.}
  \bibinfo{year}{2015}\natexlab{}.
\newblock \bibinfo{title}{{TensorFlow}: Large-Scale Machine Learning on
  Heterogeneous Systems}.
\newblock
\newblock
\urldef\tempurl%
\url{https://www.tensorflow.org/}
\showURL{%
\tempurl}
\newblock
\shownote{Software available from tensorflow.org}.


\bibitem[(ASWF)(2024)]%
        {OpenVDB}
\bibfield{author}{\bibinfo{person}{Academy Software~Foundation (ASWF)}.}
  \bibinfo{year}{2012 -- 2024}\natexlab{}.
\newblock \bibinfo{title}{{OpenVDB}}.
\newblock
\newblock
\urldef\tempurl%
\url{https://www.openvdb.org}
\showURL{%
\tempurl}


\bibitem[Barron et~al\mbox{.}(2022)]%
        {barron2022mip360}
\bibfield{author}{\bibinfo{person}{Jonathan~T Barron}, \bibinfo{person}{Ben
  Mildenhall}, \bibinfo{person}{Dor Verbin}, \bibinfo{person}{Pratul~P
  Srinivasan}, {and} \bibinfo{person}{Peter Hedman}.}
  \bibinfo{year}{2022}\natexlab{}.
\newblock \showarticletitle{Mip-nerf 360: Unbounded anti-aliased neural
  radiance fields}. In \bibinfo{booktitle}{\emph{Proceedings of the IEEE/CVF
  Conference on Computer Vision and Pattern Recognition}}.
  \bibinfo{pages}{5470--5479}.
\newblock


\bibitem[Behley et~al\mbox{.}(2021)]%
        {behley2021ijrr}
\bibfield{author}{\bibinfo{person}{J. Behley}, \bibinfo{person}{M. Garbade},
  \bibinfo{person}{A. Milioto}, \bibinfo{person}{J. Quenzel},
  \bibinfo{person}{S. Behnke}, \bibinfo{person}{J. Gall}, {and}
  \bibinfo{person}{C. Stachniss}.} \bibinfo{year}{2021}\natexlab{}.
\newblock \showarticletitle{{Towards 3D LiDAR-based semantic scene
  understanding of 3D point cloud sequences: The SemanticKITTI Dataset}}.
\newblock \bibinfo{journal}{\emph{The International Journal on Robotics
  Research}} \bibinfo{volume}{40}, \bibinfo{number}{8-9}
  (\bibinfo{year}{2021}), \bibinfo{pages}{959--967}.
\newblock
\urldef\tempurl%
\url{https://doi.org/10.1177/02783649211006735}
\showDOI{\tempurl}


\bibitem[Bradbury et~al\mbox{.}(2018)]%
        {jax2018github}
\bibfield{author}{\bibinfo{person}{James Bradbury}, \bibinfo{person}{Roy
  Frostig}, \bibinfo{person}{Peter Hawkins}, \bibinfo{person}{Matthew~James
  Johnson}, \bibinfo{person}{Chris Leary}, \bibinfo{person}{Dougal Maclaurin},
  \bibinfo{person}{George Necula}, \bibinfo{person}{Adam Paszke},
  \bibinfo{person}{Jake Vander{P}las}, \bibinfo{person}{Skye
  Wanderman-{M}ilne}, {and} \bibinfo{person}{Qiao Zhang}.}
  \bibinfo{year}{2018}\natexlab{}.
\newblock \bibinfo{booktitle}{\emph{{JAX}: composable transformations of
  {P}ython+{N}um{P}y programs}}.
\newblock
\urldef\tempurl%
\url{http://github.com/google/jax}
\showURL{%
\tempurl}


\bibitem[Chen et~al\mbox{.}(2018)]%
        {chen_brain_2018}
\bibfield{author}{\bibinfo{person}{Yuhua Chen}, \bibinfo{person}{Yibin Xie},
  \bibinfo{person}{Zhengwei Zhou}, \bibinfo{person}{Feng Shi},
  \bibinfo{person}{Anthony~G. Christodoulou}, {and} \bibinfo{person}{Debiao
  Li}.} \bibinfo{year}{2018}\natexlab{}.
\newblock \showarticletitle{Brain {MRI} super resolution using {3D} deep
  densely connected neural networks}. In \bibinfo{booktitle}{\emph{2018 {IEEE}
  15th {International} {Symposium} on {Biomedical} {Imaging} ({ISBI} 2018)}}.
  \bibinfo{publisher}{IEEE}, \bibinfo{address}{Washington, DC},
  \bibinfo{pages}{739--742}.
\newblock
\showISBNx{978-1-5386-3636-7}
\urldef\tempurl%
\url{https://doi.org/10.1109/ISBI.2018.8363679}
\showDOI{\tempurl}


\bibitem[Chollet et~al\mbox{.}(2015)]%
        {chollet2015keras}
\bibfield{author}{\bibinfo{person}{Francois Chollet} {et~al\mbox{.}}}
  \bibinfo{year}{2015}\natexlab{}.
\newblock \bibinfo{booktitle}{\emph{Keras}}.
\newblock
\urldef\tempurl%
\url{https://github.com/fchollet/keras}
\showURL{%
\tempurl}


\bibitem[Choy et~al\mbox{.}(2019)]%
        {choy20194d}
\bibfield{author}{\bibinfo{person}{Christopher Choy}, \bibinfo{person}{JunYoung
  Gwak}, {and} \bibinfo{person}{Silvio Savarese}.}
  \bibinfo{year}{2019}\natexlab{}.
\newblock \showarticletitle{4D Spatio-Temporal ConvNets: Minkowski
  Convolutional Neural Networks}. In \bibinfo{booktitle}{\emph{Proceedings of
  the IEEE Conference on Computer Vision and Pattern Recognition}}.
  \bibinfo{pages}{3075--3084}.
\newblock


\bibitem[Contributors(2022)]%
        {spconv2022}
\bibfield{author}{\bibinfo{person}{Spconv Contributors}.}
  \bibinfo{year}{2022}\natexlab{}.
\newblock \bibinfo{title}{Spconv: Spatially Sparse Convolution Library}.
\newblock \bibinfo{howpublished}{\url{https://github.com/traveller59/spconv}}.
\newblock


\bibitem[Dao et~al\mbox{.}(2022)]%
        {dao2022flashattention}
\bibfield{author}{\bibinfo{person}{Tri Dao}, \bibinfo{person}{Daniel~Y. Fu},
  \bibinfo{person}{Stefano Ermon}, \bibinfo{person}{Atri Rudra}, {and}
  \bibinfo{person}{Christopher Ré}.} \bibinfo{year}{2022}\natexlab{}.
\newblock \bibinfo{title}{FlashAttention: Fast and Memory-Efficient Exact
  Attention with IO-Awareness}.
\newblock
\newblock
\showeprint[arxiv]{2205.14135}~[cs.LG]


\bibitem[Deitke et~al\mbox{.}(2023)]%
        {Objaverse}
\bibfield{author}{\bibinfo{person}{Matt Deitke}, \bibinfo{person}{Dustin
  Schwenk}, \bibinfo{person}{Jordi Salvador}, \bibinfo{person}{Luca Weihs},
  \bibinfo{person}{Oscar Michel}, \bibinfo{person}{Eli VanderBilt},
  \bibinfo{person}{Ludwig Schmidt}, \bibinfo{person}{Kiana Ehsani},
  \bibinfo{person}{Aniruddha Kembhavi}, {and} \bibinfo{person}{Ali Farhadi}.}
  \bibinfo{year}{2023}\natexlab{}.
\newblock \showarticletitle{Objaverse: {A} Universe of Annotated 3D Objects}.
  In \bibinfo{booktitle}{\emph{Proceedings of the IEEE Conference on Computer
  Vision and Pattern Recognition (CVPR)}}. \bibinfo{pages}{13142--13153}.
\newblock


\bibitem[Geiger et~al\mbox{.}(2012)]%
        {geiger2012cvpr}
\bibfield{author}{\bibinfo{person}{A. Geiger}, \bibinfo{person}{P. Lenz}, {and}
  \bibinfo{person}{R. Urtasun}.} \bibinfo{year}{2012}\natexlab{}.
\newblock \showarticletitle{{Are we ready for Autonomous Driving? The KITTI
  Vision Benchmark Suite}}. In \bibinfo{booktitle}{\emph{Proc.~of the IEEE
  Conf.~on Computer Vision and Pattern Recognition (CVPR)}}.
  \bibinfo{pages}{3354--3361}.
\newblock


\bibitem[Hu et~al\mbox{.}(2020)]%
        {hu2019difftaichi}
\bibfield{author}{\bibinfo{person}{Yuanming Hu}, \bibinfo{person}{Luke
  Anderson}, \bibinfo{person}{Tzu-Mao Li}, \bibinfo{person}{Qi Sun},
  \bibinfo{person}{Nathan Carr}, \bibinfo{person}{Jonathan Ragan-Kelley}, {and}
  \bibinfo{person}{Fr{\'e}do Durand}.} \bibinfo{year}{2020}\natexlab{}.
\newblock \showarticletitle{DiffTaichi: Differentiable Programming for Physical
  Simulation}.
\newblock \bibinfo{journal}{\emph{ICLR}} (\bibinfo{year}{2020}).
\newblock


\bibitem[Hu et~al\mbox{.}(2019)]%
        {hu2019taichi}
\bibfield{author}{\bibinfo{person}{Yuanming Hu}, \bibinfo{person}{Tzu-Mao Li},
  \bibinfo{person}{Luke Anderson}, \bibinfo{person}{Jonathan Ragan-Kelley},
  {and} \bibinfo{person}{Fr{\'e}do Durand}.} \bibinfo{year}{2019}\natexlab{}.
\newblock \showarticletitle{Taichi: a language for high-performance computation
  on spatially sparse data structures}.
\newblock \bibinfo{journal}{\emph{ACM Transactions on Graphics (TOG)}}
  \bibinfo{volume}{38}, \bibinfo{number}{6} (\bibinfo{year}{2019}),
  \bibinfo{pages}{201}.
\newblock


\bibitem[Huang et~al\mbox{.}(2022)]%
        {10.1145/3550454.3555457}
\bibfield{author}{\bibinfo{person}{Jiahui Huang}, \bibinfo{person}{Hao-Xiang
  Chen}, {and} \bibinfo{person}{Shi-Min Hu}.} \bibinfo{year}{2022}\natexlab{}.
\newblock \showarticletitle{A Neural Galerkin Solver for Accurate Surface
  Reconstruction}.
\newblock \bibinfo{journal}{\emph{ACM Trans. Graph.}} \bibinfo{volume}{41},
  \bibinfo{number}{6}, Article \bibinfo{articleno}{229} (\bibinfo{date}{nov}
  \bibinfo{year}{2022}), \bibinfo{numpages}{16}~pages.
\newblock
\showISSN{0730-0301}
\urldef\tempurl%
\url{https://doi.org/10.1145/3550454.3555457}
\showDOI{\tempurl}


\bibitem[Huang et~al\mbox{.}(2023)]%
        {huang2023nksr}
\bibfield{author}{\bibinfo{person}{Jiahui Huang}, \bibinfo{person}{Zan Gojcic},
  \bibinfo{person}{Matan Atzmon}, \bibinfo{person}{Or Litany},
  \bibinfo{person}{Sanja Fidler}, {and} \bibinfo{person}{Francis Williams}.}
  \bibinfo{year}{2023}\natexlab{}.
\newblock \showarticletitle{Neural Kernel Surface Reconstruction}. In
  \bibinfo{booktitle}{\emph{Proceedings of the IEEE/CVF Conference on Computer
  Vision and Pattern Recognition}}. \bibinfo{pages}{4369--4379}.
\newblock


\bibitem[Jatavallabhula et~al\mbox{.}(2019)]%
        {jatavallabhula2019kaolin}
\bibfield{author}{\bibinfo{person}{Krishna~Murthy Jatavallabhula},
  \bibinfo{person}{Edward Smith}, \bibinfo{person}{Jean-Francois Lafleche},
  \bibinfo{person}{Clement~Fuji Tsang}, \bibinfo{person}{Artem Rozantsev},
  \bibinfo{person}{Wenzheng Chen}, \bibinfo{person}{Tommy Xiang},
  \bibinfo{person}{Rev Lebaredian}, {and} \bibinfo{person}{Sanja Fidler}.}
  \bibinfo{year}{2019}\natexlab{}.
\newblock \bibinfo{title}{Kaolin: A PyTorch Library for Accelerating 3D Deep
  Learning Research}.
\newblock
\newblock
\showeprint[arxiv]{1911.05063}~[cs.CV]


\bibitem[Kerbl et~al\mbox{.}(2023)]%
        {kerbl3Dgaussians}
\bibfield{author}{\bibinfo{person}{Bernhard Kerbl}, \bibinfo{person}{Georgios
  Kopanas}, \bibinfo{person}{Thomas Leimk{\"u}hler}, {and}
  \bibinfo{person}{George Drettakis}.} \bibinfo{year}{2023}\natexlab{}.
\newblock \showarticletitle{3D Gaussian Splatting for Real-Time Radiance Field
  Rendering}.
\newblock \bibinfo{journal}{\emph{ACM Transactions on Graphics}}
  \bibinfo{volume}{42}, \bibinfo{number}{4} (\bibinfo{date}{July}
  \bibinfo{year}{2023}).
\newblock
\urldef\tempurl%
\url{https://repo-sam.inria.fr/fungraph/3d-gaussian-splatting/}
\showURL{%
\tempurl}


\bibitem[Kim et~al\mbox{.}(2022)]%
        {kim2022neuralvdb}
\bibfield{author}{\bibinfo{person}{Doyub Kim}, \bibinfo{person}{Minjae Lee},
  {and} \bibinfo{person}{Ken Museth}.} \bibinfo{year}{2022}\natexlab{}.
\newblock \bibinfo{title}{NeuralVDB: High-resolution Sparse Volume
  Representation using Hierarchical Neural Networks}.
\newblock
\newblock
\showeprint[arxiv]{2208.04448}~[cs.LG]


\bibitem[Li et~al\mbox{.}(2023)]%
        {li2023nerfacc}
\bibfield{author}{\bibinfo{person}{Ruilong Li}, \bibinfo{person}{Hang Gao},
  \bibinfo{person}{Matthew Tancik}, {and} \bibinfo{person}{Angjoo Kanazawa}.}
  \bibinfo{year}{2023}\natexlab{}.
\newblock \showarticletitle{Nerfacc: Efficient sampling accelerates nerfs}.
\newblock \bibinfo{journal}{\emph{arXiv preprint arXiv:2305.04966}}
  (\bibinfo{year}{2023}).
\newblock


\bibitem[Liu et~al\mbox{.}(2023a)]%
        {liu2023one2345}
\bibfield{author}{\bibinfo{person}{Minghua Liu}, \bibinfo{person}{Ruoxi Shi},
  \bibinfo{person}{Linghao Chen}, \bibinfo{person}{Zhuoyang Zhang},
  \bibinfo{person}{Chao Xu}, \bibinfo{person}{Xinyue Wei},
  \bibinfo{person}{Hansheng Chen}, \bibinfo{person}{Chong Zeng},
  \bibinfo{person}{Jiayuan Gu}, {and} \bibinfo{person}{Hao Su}.}
  \bibinfo{year}{2023}\natexlab{a}.
\newblock \bibinfo{title}{One-2-3-45++: Fast Single Image to 3D Objects with
  Consistent Multi-View Generation and 3D Diffusion}.
\newblock
\newblock
\showeprint[arxiv]{2311.07885}~[cs.CV]


\bibitem[Liu et~al\mbox{.}(2023b)]%
        {10160968}
\bibfield{author}{\bibinfo{person}{Zhijian Liu}, \bibinfo{person}{Haotian
  Tang}, \bibinfo{person}{Alexander Amini}, \bibinfo{person}{Xinyu Yang},
  \bibinfo{person}{Huizi Mao}, \bibinfo{person}{Daniela~L. Rus}, {and}
  \bibinfo{person}{Song Han}.} \bibinfo{year}{2023}\natexlab{b}.
\newblock \showarticletitle{BEVFusion: Multi-Task Multi-Sensor Fusion with
  Unified Bird's-Eye View Representation}. In \bibinfo{booktitle}{\emph{2023
  IEEE International Conference on Robotics and Automation (ICRA)}}.
  \bibinfo{pages}{2774--2781}.
\newblock
\urldef\tempurl%
\url{https://doi.org/10.1109/ICRA48891.2023.10160968}
\showDOI{\tempurl}


\bibitem[Mane et~al\mbox{.}(2020)]%
        {mane_image_2020}
\bibfield{author}{\bibinfo{person}{Vanita Mane}, \bibinfo{person}{Suchit
  Jadhav}, {and} \bibinfo{person}{Praneya Lal}.}
  \bibinfo{year}{2020}\natexlab{}.
\newblock \showarticletitle{Image {Super}-{Resolution} for {MRI} {Images} using
  {3D} {Faster} {Super}-{Resolution} {Convolutional} {Neural} {Network}
  architecture}.
\newblock \bibinfo{journal}{\emph{ITM Web of Conferences}}
  \bibinfo{volume}{32} (\bibinfo{year}{2020}), \bibinfo{pages}{03044}.
\newblock
\showISSN{2271-2097}
\urldef\tempurl%
\url{https://doi.org/10.1051/itmconf/20203203044}
\showDOI{\tempurl}


\bibitem[Merrill(2015)]%
        {merrill2015cub}
\bibfield{author}{\bibinfo{person}{Duane Merrill}.}
  \bibinfo{year}{2015}\natexlab{}.
\newblock \showarticletitle{Cub}.
\newblock \bibinfo{journal}{\emph{NVIDIA Research}} (\bibinfo{year}{2015}).
\newblock


\bibitem[M{\"u}ller et~al\mbox{.}(2022)]%
        {muller2022instant}
\bibfield{author}{\bibinfo{person}{Thomas M{\"u}ller}, \bibinfo{person}{Alex
  Evans}, \bibinfo{person}{Christoph Schied}, {and} \bibinfo{person}{Alexander
  Keller}.} \bibinfo{year}{2022}\natexlab{}.
\newblock \showarticletitle{Instant neural graphics primitives with a
  multiresolution hash encoding}.
\newblock \bibinfo{journal}{\emph{ACM transactions on graphics (TOG)}}
  \bibinfo{volume}{41}, \bibinfo{number}{4} (\bibinfo{year}{2022}),
  \bibinfo{pages}{1--15}.
\newblock


\bibitem[Museth(2013)]%
        {Museth13:VDB}
\bibfield{author}{\bibinfo{person}{Ken Museth}.}
  \bibinfo{year}{2013}\natexlab{}.
\newblock \showarticletitle{VDB: High-Resolution Sparse Volumes with Dynamic
  Topology}.
\newblock \bibinfo{journal}{\emph{ACM Trans. Graph.}} \bibinfo{volume}{32},
  \bibinfo{number}{3}, Article \bibinfo{articleno}{27} (\bibinfo{date}{jul}
  \bibinfo{year}{2013}), \bibinfo{numpages}{22}~pages.
\newblock
\showISSN{0730-0301}
\urldef\tempurl%
\url{https://doi.org/10.1145/2487228.2487235}
\showDOI{\tempurl}


\bibitem[Museth(2014)]%
        {Museth14:HDDA}
\bibfield{author}{\bibinfo{person}{Ken Museth}.}
  \bibinfo{year}{2014}\natexlab{}.
\newblock \showarticletitle{Hierarchical Digital Differential Analyzer for
  Efficient Ray-Marching in OpenVDB}. In \bibinfo{booktitle}{\emph{ACM SIGGRAPH
  2014 Talks}} (Vancouver, Canada) \emph{(\bibinfo{series}{SIGGRAPH '14})}.
  \bibinfo{publisher}{Association for Computing Machinery},
  \bibinfo{address}{New York, NY, USA}, Article \bibinfo{articleno}{40},
  \bibinfo{numpages}{1}~pages.
\newblock
\showISBNx{9781450329606}
\urldef\tempurl%
\url{https://doi.org/10.1145/2614106.2614136}
\showDOI{\tempurl}


\bibitem[Museth(2021)]%
        {Museth21:NanoVDB}
\bibfield{author}{\bibinfo{person}{Ken Museth}.}
  \bibinfo{year}{2021}\natexlab{}.
\newblock \showarticletitle{NanoVDB: A GPU-Friendly and Portable VDB Data
  Structure For Real-Time Rendering And Simulation}. In
  \bibinfo{booktitle}{\emph{ACM SIGGRAPH 2021 Talks}} (Virtual Event, USA)
  \emph{(\bibinfo{series}{SIGGRAPH '21})}. \bibinfo{publisher}{Association for
  Computing Machinery}, \bibinfo{address}{New York, NY, USA}, Article
  \bibinfo{articleno}{1}, \bibinfo{numpages}{2}~pages.
\newblock
\showISBNx{9781450383738}
\urldef\tempurl%
\url{https://doi.org/10.1145/3450623.3464653}
\showDOI{\tempurl}


\bibitem[Paszke et~al\mbox{.}(2019)]%
        {paszke2019pytorch}
\bibfield{author}{\bibinfo{person}{Adam Paszke}, \bibinfo{person}{Sam Gross},
  \bibinfo{person}{Francisco Massa}, \bibinfo{person}{Adam Lerer},
  \bibinfo{person}{James Bradbury}, \bibinfo{person}{Gregory Chanan},
  \bibinfo{person}{Trevor Killeen}, \bibinfo{person}{Zeming Lin},
  \bibinfo{person}{Natalia Gimelshein}, \bibinfo{person}{Luca Antiga},
  \bibinfo{person}{Alban Desmaison}, \bibinfo{person}{Andreas Köpf},
  \bibinfo{person}{Edward Yang}, \bibinfo{person}{Zach DeVito},
  \bibinfo{person}{Martin Raison}, \bibinfo{person}{Alykhan Tejani},
  \bibinfo{person}{Sasank Chilamkurthy}, \bibinfo{person}{Benoit Steiner},
  \bibinfo{person}{Lu Fang}, \bibinfo{person}{Junjie Bai}, {and}
  \bibinfo{person}{Soumith Chintala}.} \bibinfo{year}{2019}\natexlab{}.
\newblock \bibinfo{title}{PyTorch: An Imperative Style, High-Performance Deep
  Learning Library}.
\newblock
\newblock
\showeprint[arxiv]{1912.01703}~[cs.LG]


\bibitem[Qi et~al\mbox{.}(2017)]%
        {qi2017pointnet}
\bibfield{author}{\bibinfo{person}{Charles~R. Qi}, \bibinfo{person}{Li Yi},
  \bibinfo{person}{Hao Su}, {and} \bibinfo{person}{Leonidas~J. Guibas}.}
  \bibinfo{year}{2017}\natexlab{}.
\newblock \bibinfo{title}{PointNet++: Deep Hierarchical Feature Learning on
  Point Sets in a Metric Space}.
\newblock
\newblock
\showeprint[arxiv]{1706.02413}~[cs.CV]


\bibitem[Ravi et~al\mbox{.}(2020)]%
        {ravi2020pytorch3d}
\bibfield{author}{\bibinfo{person}{Nikhila Ravi}, \bibinfo{person}{Jeremy
  Reizenstein}, \bibinfo{person}{David Novotny}, \bibinfo{person}{Taylor
  Gordon}, \bibinfo{person}{Wan-Yen Lo}, \bibinfo{person}{Justin Johnson},
  {and} \bibinfo{person}{Georgia Gkioxari}.} \bibinfo{year}{2020}\natexlab{}.
\newblock \showarticletitle{Accelerating 3D Deep Learning with PyTorch3D}.
\newblock \bibinfo{journal}{\emph{arXiv:2007.08501}} (\bibinfo{year}{2020}).
\newblock


\bibitem[Ren et~al\mbox{.}(2023)]%
        {ren2023xcube}
\bibfield{author}{\bibinfo{person}{Xuanchi Ren}, \bibinfo{person}{Jiahui
  Huang}, \bibinfo{person}{Xiaohui Zeng}, \bibinfo{person}{Ken Museth},
  \bibinfo{person}{Sanja Fidler}, {and} \bibinfo{person}{Francis Williams}.}
  \bibinfo{year}{2023}\natexlab{}.
\newblock \showarticletitle{XCube: Large-Scale 3D Generative Modeling using
  Sparse Voxel Hierarchies}.
\newblock \bibinfo{journal}{\emph{arXiv preprint}} (\bibinfo{year}{2023}).
\newblock


\bibitem[Shi et~al\mbox{.}(2020)]%
        {Shi_2020_CVPR}
\bibfield{author}{\bibinfo{person}{Shaoshuai Shi}, \bibinfo{person}{Chaoxu
  Guo}, \bibinfo{person}{Li Jiang}, \bibinfo{person}{Zhe Wang},
  \bibinfo{person}{Jianping Shi}, \bibinfo{person}{Xiaogang Wang}, {and}
  \bibinfo{person}{Hongsheng Li}.} \bibinfo{year}{2020}\natexlab{}.
\newblock \showarticletitle{PV-RCNN: Point-Voxel Feature Set Abstraction for 3D
  Object Detection}. In \bibinfo{booktitle}{\emph{Proceedings of the IEEE/CVF
  Conference on Computer Vision and Pattern Recognition (CVPR)}}.
\newblock


\bibitem[Sun et~al\mbox{.}(2020)]%
        {waymo}
\bibfield{author}{\bibinfo{person}{Pei Sun}, \bibinfo{person}{Henrik
  Kretzschmar}, \bibinfo{person}{Xerxes Dotiwalla}, \bibinfo{person}{Aurelien
  Chouard}, \bibinfo{person}{Vijaysai Patnaik}, \bibinfo{person}{Paul Tsui},
  \bibinfo{person}{James Guo}, \bibinfo{person}{Yin Zhou},
  \bibinfo{person}{Yuning Chai}, \bibinfo{person}{Benjamin Caine},
  \bibinfo{person}{Vijay Vasudevan}, \bibinfo{person}{Wei Han},
  \bibinfo{person}{Jiquan Ngiam}, \bibinfo{person}{Hang Zhao},
  \bibinfo{person}{Aleksei Timofeev}, \bibinfo{person}{Scott Ettinger},
  \bibinfo{person}{Maxim Krivokon}, \bibinfo{person}{Amy Gao},
  \bibinfo{person}{Aditya Joshi}, \bibinfo{person}{Yu Zhang},
  \bibinfo{person}{Jonathon Shlens}, \bibinfo{person}{Zhifeng Chen}, {and}
  \bibinfo{person}{Dragomir Anguelov}.} \bibinfo{year}{2020}\natexlab{}.
\newblock \showarticletitle{Scalability in Perception for Autonomous Driving:
  Waymo Open Dataset}. In \bibinfo{booktitle}{\emph{Proceedings of the IEEE
  Conference on Computer Vision and Pattern Recognition (CVPR)}}.
  \bibinfo{pages}{2443--2451}.
\newblock


\bibitem[Tancik et~al\mbox{.}(2023)]%
        {nerfstudio}
\bibfield{author}{\bibinfo{person}{Matthew Tancik}, \bibinfo{person}{Ethan
  Weber}, \bibinfo{person}{Evonne Ng}, \bibinfo{person}{Ruilong Li},
  \bibinfo{person}{Brent Yi}, \bibinfo{person}{Justin Kerr},
  \bibinfo{person}{Terrance Wang}, \bibinfo{person}{Alexander Kristoffersen},
  \bibinfo{person}{Jake Austin}, \bibinfo{person}{Kamyar Salahi},
  \bibinfo{person}{Abhik Ahuja}, \bibinfo{person}{David McAllister}, {and}
  \bibinfo{person}{Angjoo Kanazawa}.} \bibinfo{year}{2023}\natexlab{}.
\newblock \showarticletitle{Nerfstudio: A Modular Framework for Neural Radiance
  Field Development}. In \bibinfo{booktitle}{\emph{ACM SIGGRAPH 2023 Conference
  Proceedings}} \emph{(\bibinfo{series}{SIGGRAPH '23})}.
\newblock


\bibitem[Tang et~al\mbox{.}(2022)]%
        {tang2022torchsparse}
\bibfield{author}{\bibinfo{person}{Haotian Tang}, \bibinfo{person}{Zhijian
  Liu}, \bibinfo{person}{Xiuyu Li}, \bibinfo{person}{Yujun Lin}, {and}
  \bibinfo{person}{Song Han}.} \bibinfo{year}{2022}\natexlab{}.
\newblock \showarticletitle{{TorchSparse: Efficient Point Cloud Inference
  Engine}}. In \bibinfo{booktitle}{\emph{Conference on Machine Learning and
  Systems (MLSys)}}. \bibinfo{address}{Indio, CA, USA}.
\newblock


\bibitem[Tang et~al\mbox{.}(2023)]%
        {tangandyang2023torchsparse}
\bibfield{author}{\bibinfo{person}{Haotian Tang}, \bibinfo{person}{Shang Yang},
  \bibinfo{person}{Zhijian Liu}, \bibinfo{person}{Ke Hong},
  \bibinfo{person}{Zhongming Yu}, \bibinfo{person}{Xiuyu Li},
  \bibinfo{person}{Guohao Dai}, \bibinfo{person}{Yu Wang}, {and}
  \bibinfo{person}{Song Han}.} \bibinfo{year}{2023}\natexlab{}.
\newblock \showarticletitle{TorchSparse++: Efficient Training and Inference
  Framework for Sparse Convolution on GPUs}. In
  \bibinfo{booktitle}{\emph{IEEE/ACM International Symposium on
  Microarchitecture (MICRO)}}.
\newblock


\bibitem[Thakkar et~al\mbox{.}(2023)]%
        {Thakkar_CUTLASS_2023}
\bibfield{author}{\bibinfo{person}{Vijay Thakkar}, \bibinfo{person}{Pradeep
  Ramani}, \bibinfo{person}{Cris Cecka}, \bibinfo{person}{Aniket Shivam},
  \bibinfo{person}{Honghao Lu}, \bibinfo{person}{Ethan Yan},
  \bibinfo{person}{Jack Kosaian}, \bibinfo{person}{Mark Hoemmen},
  \bibinfo{person}{Haicheng Wu}, \bibinfo{person}{Andrew Kerr},
  \bibinfo{person}{Matt Nicely}, \bibinfo{person}{Duane Merrill},
  \bibinfo{person}{Dustyn Blasig}, \bibinfo{person}{Fengqi Qiao},
  \bibinfo{person}{Piotr Majcher}, \bibinfo{person}{Paul Springer},
  \bibinfo{person}{Markus Hohnerbach}, \bibinfo{person}{Jin Wang}, {and}
  \bibinfo{person}{Manish Gupta}.} \bibinfo{year}{2023}\natexlab{}.
\newblock \bibinfo{booktitle}{\emph{{CUTLASS}}}.
\newblock
\urldef\tempurl%
\url{https://github.com/NVIDIA/cutlass}
\showURL{%
\tempurl}


\bibitem[Wang et~al\mbox{.}(2017)]%
        {Wang_2017}
\bibfield{author}{\bibinfo{person}{Peng-Shuai Wang}, \bibinfo{person}{Yang
  Liu}, \bibinfo{person}{Yu-Xiao Guo}, \bibinfo{person}{Chun-Yu Sun}, {and}
  \bibinfo{person}{Xin Tong}.} \bibinfo{year}{2017}\natexlab{}.
\newblock \showarticletitle{O-CNN: octree-based convolutional neural networks
  for 3D shape analysis}.
\newblock \bibinfo{journal}{\emph{ACM Transactions on Graphics}}
  \bibinfo{volume}{36}, \bibinfo{number}{4} (\bibinfo{date}{July}
  \bibinfo{year}{2017}), \bibinfo{pages}{1–11}.
\newblock
\showISSN{1557-7368}
\urldef\tempurl%
\url{https://doi.org/10.1145/3072959.3073608}
\showDOI{\tempurl}


\bibitem[Zhao et~al\mbox{.}(2021)]%
        {Zhao_2021_ICCV}
\bibfield{author}{\bibinfo{person}{Hengshuang Zhao}, \bibinfo{person}{Li
  Jiang}, \bibinfo{person}{Jiaya Jia}, \bibinfo{person}{Philip~H.S. Torr},
  {and} \bibinfo{person}{Vladlen Koltun}.} \bibinfo{year}{2021}\natexlab{}.
\newblock \showarticletitle{Point Transformer}. In
  \bibinfo{booktitle}{\emph{Proceedings of the IEEE/CVF International
  Conference on Computer Vision (ICCV)}}. \bibinfo{address}{New York, NY, USA},
  \bibinfo{pages}{16259--16268}.
\newblock


\end{thebibliography}

\end{document}